\pdfoutput=1
\documentclass{article}
\usepackage{aaai25}  
\usepackage{times}  
\usepackage{helvet}  
\usepackage{courier}  
\usepackage[hyphens]{url}  
\usepackage{graphicx} 
\usepackage{natbib}  
\usepackage{caption} 
\usepackage{algorithm}
\usepackage{algorithmic}
\usepackage{newfloat}
\usepackage{listings}
\DeclareCaptionStyle{ruled}{labelfont=normalfont,labelsep=colon,strut=off} 
\floatstyle{ruled}
\newfloat{listing}{tb}{lst}{}
\floatname{listing}{Listing}
\nocopyright

\title{ArtiFade: Learning to Generate High-quality Subject from Blemished Images}
\author{
    Shuya Yang\thanks{Equal contribution}, 
    Shaozhe Hao\footnotemark[1], 
    Yukang Cao\thanks{Corresponding authors}, 
    Kwan-Yee K. Wong\footnotemark[2]
}
\affiliations{
    The University of Hong Kong
}

\usepackage{bibentry}

\usepackage[dvipsnames]{xcolor}
\usepackage{amsmath}
\usepackage{xspace}
\usepackage{eccvabbrv_template}
\usepackage{multirow}
\usepackage{booktabs}
\usepackage{array}
\usepackage{tikz}
\usepackage{amssymb}
\usepackage{caption}
\usepackage{subcaption}
\usepackage{cleveref}
\usepackage{graphicx}

\begin{document}
\newcommand{\udensdot}[1]{%
    \tikz[baseline=(todotted.base)]{
        \node[inner sep=1pt,outer sep=0pt] (todotted) {#1};
        \draw[dotted, thick] (todotted.south west) -- (todotted.south east);
    }%
}%
\definecolor{blue-violet}{rgb}{0.54, 0.17, 0.89}
\newcommand{\yk}[1]{\textcolor{blue-violet}{#1}}
\newcommand{\hsz}[1]{\textcolor{black}{#1}}
\newcommand{\TBD}[1]{\textcolor{red}{[#1]}}
\newcommand{\NEW}[1]{\textcolor{YellowOrange}{#1}}
\newcommand*\samethanks[1][\value{footnote}]{\footnotemark[#1]}
\def\VName{ArtiFade\xspace}
\def\VEmbed{\textlangle$\mathrm{\Phi}$\textrangle\xspace}
\def\WMModel{\texttt{\udensdot{{WM-model}}}\xspace}
\def\RCModel{\texttt{\underline{RC-model}}\xspace}
\def\ANModel{\texttt{\textcolor{OliveGreen}{AN-model}}\xspace}
\newcommand{\sy}[1]{\textcolor{black}{#1}}
\maketitle
\begin{abstract}
Subject-driven text-to-image generation has witnessed remarkable advancements in its ability to learn and capture characteristics of a subject using only a limited number of images. However, existing methods commonly rely on high-quality images for training and may struggle to generate reasonable images when the input images are blemished by artifacts. This is primarily attributed to the inadequate capability of current techniques in distinguishing subject-related features from disruptive artifacts. In this paper, we introduce ArtiFade to tackle this issue and successfully generate high-quality artifact-free images from blemished datasets. Specifically, ArtiFade exploits fine-tuning of a pre-trained text-to-image model, aiming to remove artifacts. The elimination of artifacts is achieved by utilizing a specialized dataset that encompasses both unblemished images and their corresponding blemished counterparts during fine-tuning. ArtiFade also ensures the preservation of the original generative capabilities inherent within the diffusion model, thereby enhancing the overall performance of subject-driven methods in generating high-quality and artifact-free images. We further devise evaluation benchmarks tailored for this task. Through extensive qualitative and quantitative experiments, we demonstrate the generalizability of ArtiFade in effective artifact removal under both in-distribution and out-of-distribution scenarios. 
\end{abstract}

\section{Introduction}
\label{sec:intro}
\begin{figure*}[t]
\includegraphics[width=1.0\linewidth]{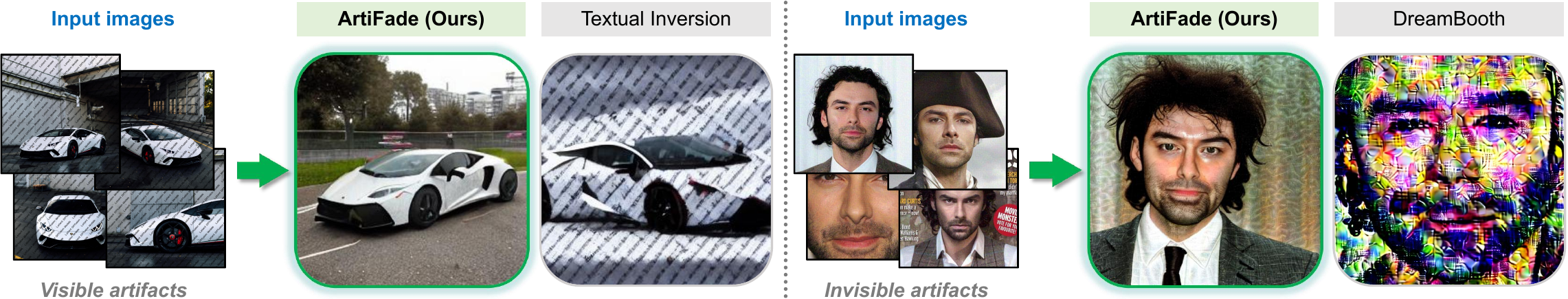}
\caption{\hsz{Blemished subject-driven generation with our \VName and vanilla subject-driven methods.
We display images generated using \VName and Textual Inversion on watermark artifacts on the left, and \VName and DreamBooth on adversarial noise artifacts~\cite{van2023anti} on the right. In contrast to the poor performance of Textual Inversion and DreamBooth, which are negatively affected by the visiable or invisible artifacts, ArtiFade produces much better fidelity of the subject with high-quality generation.}}
\label{fig:teaser}
\end{figure*}
With the rapid advancement of generative diffusion models~\cite{rombach2022high, song2021denoising, saharia2022photorealistic, Zhang_2023_ICCV, ho2020denoising}, subject-driven text-to-image generation~\cite{gal2022image,ruiz2023dreambooth, kumari2023multi, Kawar_2023_CVPR, NEURIPS2023_6091bf15}, which aims to capture distinct characteristics of a subject by learning from a few images of the subject, has gained significant attention.
This approach empowers individuals to seamlessly incorporate their preferred subjects into diverse and visually captivating scenes by simply providing text conditions.
Representative works such as Textual Inversion~\cite{gal2022image} and DreamBooth~\cite{ruiz2023dreambooth} have shown promising results on this task. Specifically, Textual Inversion proposes to optimize a textual embedding to encode identity characteristics that provide rich subject information for subsequent generation. DreamBooth shares a similar idea but additionally fine-tunes the diffusion model to preserve more identity semantics. Plenty of successive efforts have been made to advance this task from various perspectives, including generation quality, compositionality, and efficiency~\cite{kumari2023multi, NEURIPS2023_6091bf15, Kawar_2023_CVPR}.

Both of the above mentioned methods, along with their follow-up works, however, rely heavily on the presence of unblemished input images that contain only relevant identity information. This is often expensive or even unavailable in real-world applications. 
Instead, in practical scenarios such as scraping web images of a desired subject, it is common to encounter images that are blemished by various \emph{visible} artifacts such as watermarks, drawings, and stickers.
Additionally, there also exist \emph{invisible} artifacts like adversarial noises~\cite{van2023anti} that are not easily detectable or removable using off-the-shelf tools. These artifacts can significantly impede the comprehensive learning of the subject and lead to a catastrophic decline in performance across multiple dimensions (see Fig. \ref{fig:teaser}). This limitation arises from the feature confusion inherent in the existing subject-driven learning process. The process simultaneously captures subject-related features and disruptive artifact interference. It lacks the discriminative power to distinguish these two from each other, and fails to preserve the integrity of subject characteristics while mitigating negative effects caused by artifacts. 
As blemished inputs are inevitable in applications, a pressing challenge emerges: \textbf{Can we effectively perform subject-driven text-to-image generation using \emph{blemished} images?} We term this novel problem (\ie, generating subject-driven images from blemished inputs) as blemished subject-driven generation in this paper.

To answer the above question, we present \textbf{ArtiFade}, the first model to tackle blemished subject-driven generation by adapting vanilla subject-driven methods (\eg, Textual Inversion~\cite{gal2022image} and DreamBooth~\cite{ruiz2023dreambooth}) to effectively extract subject-specific information from blemished training data. The key objective of ArtiFade is to learn the implicit relationship between natural images and their blemished counterparts through alignment optimization.
Specifically, we introduce a specialized dataset construction method to create pairs of unblemished images and their corresponding counterparts. These pairs can be applied to fine-tune various subject-driven approaches in the context of blemished subject-driven generation. Besides, we also observe fine-tuning an extra learnable embedding in the textual space, named artifact-free embedding, can enhance prompt fidelity in the blemished subject-driven generation.

We further introduce an evaluation benchmark that encompasses \textbf{(1)} multiple test sets of blemished images with diverse artifacts, and \textbf{(2)} tailored metrics for accurately assessing the performance of blemished subject-driven generation methods. A thorough experimental evaluation shows that our method consistently outperforms other existing methods, both qualitatively and quantitatively. Notably, \VName exhibits superb capabilities in handling out-of-distribution (OOD) scenarios involving diverse types of artifacts that are distinct from the training data. This inherent generalizability indicates our model can effectively learn to discern and distinguish the patterns exhibited by artifacts and unblemished images, instead of overfitting to a specific type of artifacts.

In summary, our key contributions are as follows:
\begin{itemize}
  \item We are the first to tackle the novel challenge of blemished subject-driven generation. To address this task, we propose \VName that fine-tunes diffusion models to align unblemished and blemished data.
  \item We introduce an evaluation benchmark tailored for effectively assessing the performance of blemished subject-driven generation techniques.
  \item We conduct extensive experiments and demonstrate that \VName outperforms current methods significantly. We show noteworthy generalizability of \VName, effectively addressing both in-distribution and out-of-distribution scenarios with various types of artifacts.
\end{itemize}
\section{Related Work}
\label{sec:relatedWork}

\begin{figure*}[t]
  \centering
  \includegraphics[width=1.0\linewidth, trim=4 4 4 4]{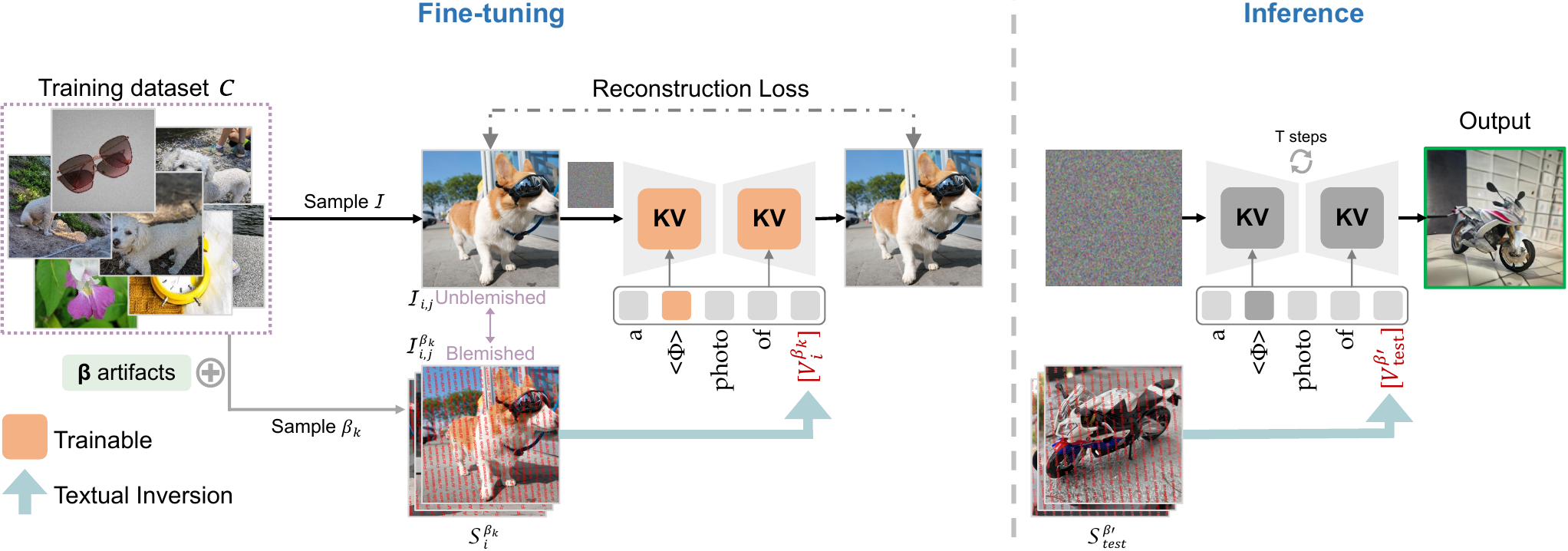}
  \caption{Overview of \VName. On the left, we present Artifact Rectification Training, which involves an iterative process of calculating reconstruction loss between an unblemished image and the reconstruction of its blemished embedding. The right-hand side is the inference stage that tests \VName on unseen blemished images. To avoid ambiguity, we (1) simplify the training of Textual Inversion into an input-output form, and (2) use ``fine-tuning'' and ``inference'' to respectively refer to the fine-tuning stage of \VName and the use of \VName for subject-driven generation.}
  \label{fig:flow}
\end{figure*}

\paragraph{\bf Text-to-image synthesis} 
Text-to-image generation has attracted considerable attention in recent years by leveraging Generative Adversarial Networks (GANs)~\cite{NIPS2014_5ca3e9b1} and diffusion models~\cite{ho2020denoising, rombach2022high}. Reed~\etal~\shortcite{reed2016generative} was the first to integrate GANs into text-to-image generation. Since then, several influential works had been proposed~\cite{zhang2017stackgan, zhang2018stackgan++, xu2018attngan, zhang2018photographic, zhu2019dm, li2019controllable, ruan2021dae, zhang2021cross, cheng2020rifegan, qiao2019mirrorgan, yin2019semantics}, demonstrating impressive results with improved resolution~\cite{zhang2017stackgan, zhang2018stackgan++} and fidelity of fine details~\cite{xu2018attngan}. Diffusion models in text-to-image synthesis have also yielded remarkable results owing to their ability in generating precise and customized images that better align with individual text specifications~\cite{nichol2021glide, saharia2022photorealistic, ramesh2022hierarchical, gu2022vector, rombach2022high}. 

\paragraph{\bf Subject-driven generation} 
Subject-driven generation has gained popularity due to its ability to generate personalized images based on a given set of subject images and text prompts. One prominent method in subject-driven generation is Textual Inversion~\cite{gal2022image}, which involves learning an embedding vector by minimizing the Latent Diffusion Model loss~\cite{rombach2022high} on input images. The learned embedding vector can be effectively combined with text prompts, allowing seamless integration in the text-to-image generation process. Recent approaches~\cite{ruiz2023dreambooth, kumari2023multi, Lu_2023_CVPR} have significantly enhanced subject reconstruction fidelity by incorporating fine-tuning techniques.

\paragraph{\bf Artifacts removal}
Shadow and watermark removal are classic tasks in image processing and computer vision. At the early stage, most approaches for shadow removal or image recovery relied on the properties of intensity and illumination~\cite{finlayson2009entropy, 1542031, 7180373, xiao2013efficient, xiao2013fast, finlayson2002removing, khan2015automatic, shor2008shadow, arbel2010shadow, guo2011single}. Some methods also incorporated color features to improve their results~\cite{guo2011single}. Deep learning techniques and Convolutional Neural Networks (CNNs) have played a significant role in advancing shadow removal methods and producing impressive results in recent years~\cite{ding2019argan, hu2019mask, le2019shadow, liu2021shadow, wang2018stacked, Zhu_2022_CVPR, chen2021canet, jin2023estimating, fu2021auto}. Several studies~\cite{wang2018stacked, liu2021shadow, hu2019mask, ding2019argan} have incorporated GANs to further enhance the results of shadow removal techniques. Moreover, with the increasing popularity of diffusion models in image generation, a novel diffusion-based method for shadow removal has recently been introduced~\cite{guo2023shadowdiffusion}.

The most widely adopted methods for recovering concealed information from watermarked images include the application of generalized multi-image matting algorithms~\cite{dekel2017effectiveness}, complemented by image inpainting techniques~\cite{xu2017automatic, qin2018visible, huang2004attacking}, and the utilization of deep neural networks and CNNs~\cite{cheng2018large}. Similar to shadow removal, GANs and Conditional GANs~\cite{mirza2014conditional} are also widely used in watermark removal tasks~\cite{li2019towards, cao2019generative, Liu_2021_WACV}. Our work is closely related to these previously mentioned studies. We are the first to address the artifact issues in the realm of subject-driven text-to-image generation.
\section{Method}
\label{sec:method} 
Given a set of blemished input images, our objective is to eliminate their negative impacts on the quality of subject-driven image generation.
To achieve this goal, we present \VName, an efficient framework that learns to discern and distinguish the patterns exhibited by various types of artifacts and unblemished images.
In this section, we focus exclusively on ArtiFade based on Textual Inversion. However, it is important to note that the ArtiFade framework can be generalized to other subject-driven generation methods.
As shown in Fig. \ref{fig:flow}, \VName based on Textual Inversion incorporates two main components, namely the fine-tuning of the partial parameters (\ie, key and value weights) in the diffusion model and the simultaneous optimization of an artifact-free embedding \VEmbed. We begin by discussing the preliminaries of the Latent Diffusion Model and Textual Inversion. In the following subsections, we elaborate our automatic construction of the training dataset, which consists of both blemished and unblemished data, illustrated in Sec. \ref{subsec:dataset}. 
\hsz{We then introduce Artifact Rectification Training, a method for fine-tuning the model to accommodate blemished images, as discussed in Sec. \ref{subsec: workflow}}.
We finally present the use of \VName for handling blemished images in Sec. \ref{subsec:inference}. 
\paragraph{{\bf Preliminary}} Latent Diffusion Model (LDM)~\cite{rombach2022high} is a latent text-to-image diffusion model derived from Diffusion Denoising Probabilistic Model (DDPM)~\cite{ho2020denoising}. 
LDM leverages a pre-trained autoencoder to map image features between the image and latent space. This autoencoder comprises an encoder $\mathcal{E}$, which transforms images into latent representations, and a decoder $\mathcal{D}$, which converts latent representations back into images. The autoencoder is optimized using a set of images so that the reconstructed image $\hat{x} \approx \mathcal{D}(\mathcal{E}(x))$. Additionally, LDM introduces cross-attention layers~\cite{vaswani2017attention} within the U-Net~\cite{ronneberger2015u}, enabling the integration of text prompts as conditional information during the image generation process.
The LDM loss is defined as
\begin{equation}
  \mathcal{L}_{LDM} := \mathbb{E}_{z \sim \mathcal{E}(\mathcal{I}), y, \epsilon \sim N(0,1)}
  \Bigl[\lVert\epsilon - \epsilon_{\theta}(z_{t}, t, y) \rVert_{2}^{2}\Bigr],
  \label{eq:LDMloss}
\end{equation} 
where $\mathcal{E}$ encodes the image $\mathcal{I}$ into the latent representation $z$. Here, $z_{t}$ denotes the noisy latent representation at timestep $t$, $\epsilon_{\theta}$ refers to the denoising network, and $y$ represents the text condition that is passed to the cross-attention layer.

Based on LDM, Textual Inversion~\cite{gal2022image} aims to capture the characteristics of a specific subject from a small set of images. Specifically, Textual Inversion learns a unique textual embedding by minimizing Eq.~(\ref{eq:LDMloss}) on a few images that contain the particular subject. It can produce promising generation results with high-quality inputs, but fails on input images that are blemished by artifacts (see Fig.~\ref{fig:teaser}). This problem arises from the inherent limitation of Textual Inversion in learning shared characteristics exhibited in the input images without the capability in differentiating artifacts from unblemished subjects. In this paper, we aim to address this issue on deteriorated generation quality of Textual Inversion in the presence of blemished images.

\subsection{Dataset Construction}
\label{subsec:dataset}
Existing subject-driven generation methods operate under the assumption of unblemished training data, consisting of solely high-quality images devoid of any artifacts. However, this assumption does not align with real-world applications, where obtaining blemished images from the internet is a commonplace. To address this blemished subject-driven generation in this paper, we first construct a training set that incorporates both unblemished images and their blemished counterparts that are augmented with artifacts.

\paragraph{\bf Augmentation of multiple artifacts} 
We construct our dataset by collecting a multi-subject set $\mathcal{C}$ of $N$ image subsets from existing works~\cite{gal2022image,ruiz2023dreambooth, kumari2023multi} and a set $\mathcal{B}$ of $L$ different artifacts:
\begin{equation}
\label{eq:setting}
    \mathcal{C} = \{\mathcal{S}_i\}_{i=1}^N, \quad \mathcal{S}_i = \{\mathcal{I}_{i,j}\}_{j=1}^{M_i}, \quad \mathcal{B}=\{\beta_k\}_{k=1}^L,
\end{equation}
where $\mathcal{S}_i$ denotes the image subset corresponding to the $i$th subject, $M_i$ is the total number of images in $\mathcal{S}_i$, and $\beta_k$ represents a type of artifact for image augmentation. Our dataset $\mathcal{D}$ can then be constructed by applying each artifact $\beta_k$ to each image $\mathcal{I}$ in $\mathcal{S}_i$ separately, i.e.,
\begin{equation}
 \mathcal{S}_i^{\beta_k} = \{\mathcal{I}_{i,j}^{\beta_k}\}_{j=1}^{M_i}, \quad \mathcal{D} = \{\mathcal{S}_i, \{\mathcal{S}_i^{\beta_k}\}_{k=1}^L\}_{i=1}^N,
\end{equation}
where $\mathcal{I}_{i,j}^{\beta_k}$ is the counterpart of $\mathcal{I}_{i,j}$ augmented with the specific artifact $\beta_k$. Some examples of original images and their augmented versions with distinct artifacts can be found in Fig.~\ref{fig:trainingDataset}. See the Appendix for more visualizations.

\begin{figure}[t]
  \centering
  \includegraphics[width=1.0\linewidth, trim=4 4 4 4]{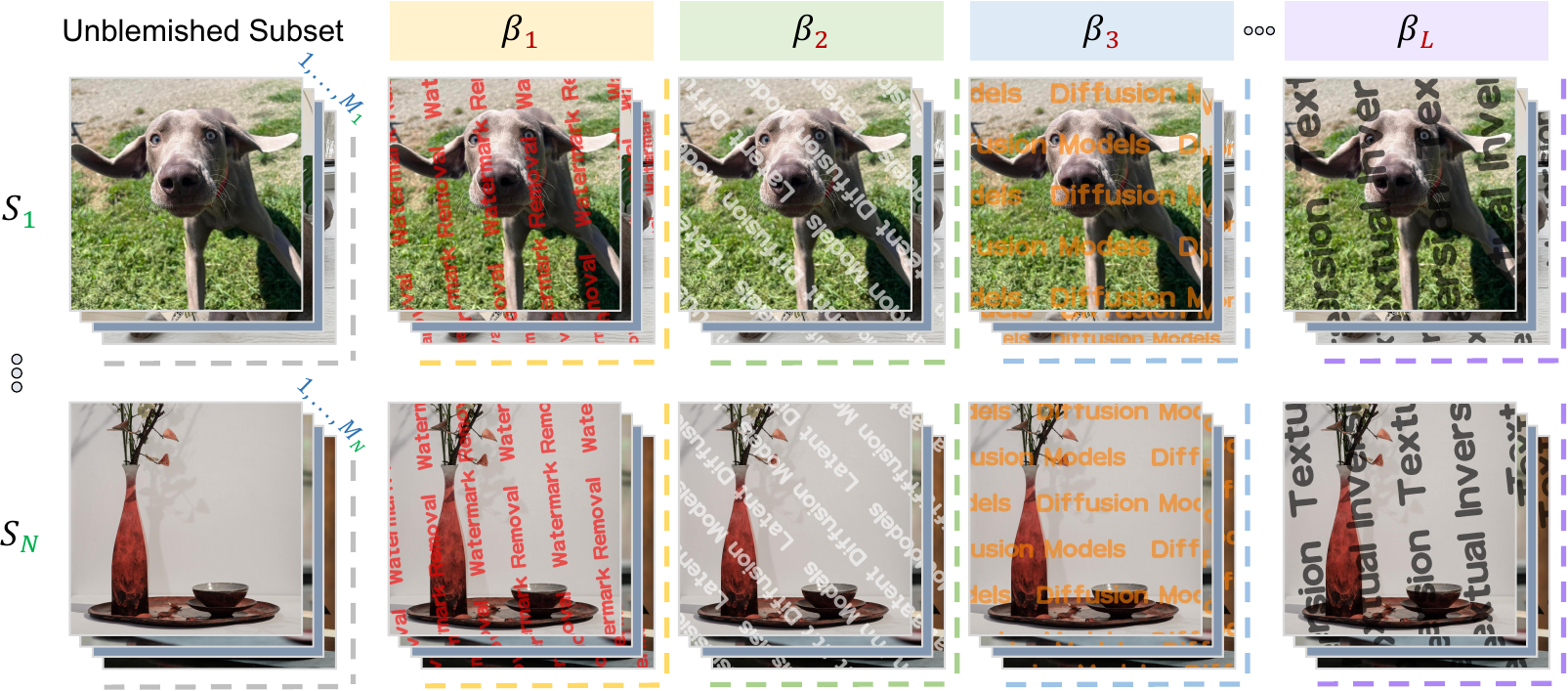}
  \caption{Examples of training dataset $\mathcal{D}$ that contains both unblemished images and blemished counterparts.}
  \label{fig:trainingDataset}
\end{figure}
\paragraph{\bf Blemished textual embedding} 

For each blemished subset, we perform Textual Inversion to optimize a blemished textual embedding $[\mathtt{V}_i^{\beta_k}]$ 
, i.e.,
\begin{equation} 
\label{eq:bti}
    \begin{aligned}
        \mathcal{S}_i^{\beta_k} \quad \underrightarrow{\text{Textual Inversion}} \quad [\mathtt{V}_i^{\beta_k}],\\ \quad i=1,2,...,N;\ \ k=1,2,...,L
    \end{aligned}
\end{equation}
By applying Eq.~(\ref{eq:bti}) on $N$ subsets with $L$ types of artifacts, we end up with a set of $N \times L$ blemished textual embeddings $\mathcal{V} = \{[\mathtt{V}_i^{\beta_k}]\}_{i=1,k=1}^{N,L}$, which will be used in the subsequent model fine-tuning. As we have illustrated in Fig.~\ref{fig:teaser}, directly prompting the diffusion model with $[\mathtt{V}_i^{\beta_k}]$ will lead to a significant decrease in generation quality. Consequently, our objective is to robustly handle blemished embeddings and effectively eliminate the detrimental impact of artifacts. We achieve this by devising a partial fine-tuning paradigm for the pre-trained diffusion model on the constructed training set $\mathcal{D}$, as elaborated in the following subsection.

\subsection{Artifact Rectification Training} 
\label{subsec: workflow}
After establishing the curated dataset $\mathcal{D}$, we embark on training a generalizable framework on $\mathcal{D}$, capable of generating unblemished images using blemished textual embeddings.
To this end, we propose artifact rectification training, which consists of two key components, namely partial fine-tuning of a pre-trained diffusion model and the optimization of an artifact-free embedding, to eliminate the artifacts and distortions in the generated images.

We fine-tune only partial parameters that are involved in processing the textual conditions. This strategy allows us to optimize the relevant components associated with the blemished textual embedding $[\mathtt{V}_i^{\beta_k}]$. Considering that only the key and value weights in the diffusion model's cross-attention layer are involved in the processing of textual embedding, we choose to fine-tune these two types of parameters $W^{k}$ and $W^{v}$. Moreover, we find that optimizing an additional embedding, \VEmbed, in the textual space with partial parameters could improve prompt fidelity by retaining the textual information of the model, as presented later in Sec. \ref{subsec: AblationStudy}.

\paragraph{\bf Training objective}
During each iteration, we will first randomly sample an unblemished image $\mathcal{I}_{i,j}$ from the training set $\mathcal{D}$ and a type of artifact $\beta_k \in \mathcal{B}$ to obtain
the blemished textual embedding $[\mathtt{V}_i^{\beta_k}] \in \mathcal{V}$ that is optimized on the blemished subset $\mathcal{S}_i^{\beta_k}$.

Specifically, given the sampled blemished textual embedding $[\mathtt{V}_i^{\beta_k}]$, we form the prompt ``a \VEmbed photo of $[\mathtt{V}_i^{\beta_k}]$'', which will be input to the text encoder to acquire the text condition $y_i^{\beta_k}$. 
Our optimization objective will then be defined as reconstructing the unblemished image $\mathcal{I}_{i,j}$ by conditioning the denoising process on the text condition $y_i^{\beta_k}$.
Thus, we can formulate the final loss for training \VName as 
\begin{equation}
    \begin{aligned}
    \mathcal{L}_\text{ArtiFade} :=& \mathbb{E}_{z \sim \mathcal{E}(\mathcal{{I}}_{i,j}), y_i^{\beta_k}, \epsilon \sim N(0,1)} \\
  &\Bigl[\lVert\epsilon - \epsilon_{\{W^k, W^v, \langle {\mathrm{\Phi}} \rangle \}}(z_{t}, t, {y_i^{\beta_k}})\rVert_{2}^{2}\Bigr],      
    \end{aligned}
  \label{eq:ArtiFade loss}
\end{equation}
where $\{W^k, W^v, \langle {\mathrm{\Phi}} \rangle \}$ is the set of the trainable parameters of \VName.

\subsection{Subject-driven Generation with Blemished Images}
\label{subsec:inference}

After artifact rectification training, we obtain the ArtiFade model, prepared for the task of blemished subject-driven generation.
Given a test image set $\mathcal{S}^{\beta^
\prime}_{\text{test}}$ in which all images are blemished by an arbitrary artifact $\beta^\prime$,
the \VName model can generate high-quality subject-driven images using blemished samples with ease.

Specifically, we first obtain the blemished textual embedding $\mathtt{[V_{\text{test}}^{\beta^\prime}]}$ by applying Textual Inversion on the test set $\mathcal{S}^{\beta^
\prime}_{\text{test}}$. We then simply infer the \VName model with a given text prompt that includes the blemished textual embedding, \ie, ``a \VEmbed photo of $\mathtt{[V_{\text{test}}^{\beta^\prime}]}$''. At the operational level, the sole distinction between our approach and vanilla Textual Inversion lies in inputting text prompts containing $\mathtt{[V_{\text{test}}^{\beta^\prime}]}$ into the fine-tuned \VName instead of the pre-trained diffusion model. This simple yet effective method resolves the issue of Textual Inversion's incapacity to handle blemished input images, bearing practical utility.

\paragraph{{\bf Details of ArtiFade models}} 
We choose $N=$ 20 subjects, including pets, plants, containers, toys, and wearable items to ensure a diverse range of categories.
We experiment with the ArtiFade model based on Textual Inversion trained with visible watermark artifacts, namely \WMModel.
The training set of \WMModel involves $L_{\text{WM}}=$ 10 types of watermarks, characterized by various fonts, orientations, colors, sizes, and text contents. Therefore, we obtain 200 blemished subsets in total within the training set of \WMModel. We fine-tune \WMModel for a total of 16k steps.
\begin{table}[t]
  \centering
\fontsize{9pt}{9pt}\selectfont
\setlength{\tabcolsep}{3pt}
\begin{tabular}{l @{\hskip 0.1in} ccccc @{\hskip 0.1in} ccccc}
    \toprule
    \multirow{2}{*}{Method} &     \multicolumn{5}{c}{\texttt{\WMModel on WM-ID-test}} \\
      \cmidrule(lr{0.1in}){2-6}
      & {I\textsuperscript{DINO}$\uparrow$} & {R\textsuperscript{DINO}$\uparrow$} & {I\textsuperscript{CLIP}$\uparrow$} & {R\textsuperscript{CLIP}$\uparrow$} & {T\textsuperscript{CLIP}$\uparrow$} \\
      
      \midrule
    \color{lightgray}TI (unblemished) 
    & \color{lightgray}0.488 & \color{lightgray}1.349 & \color{lightgray}0.730 & \color{lightgray}1.070 & \color{lightgray}0.283 \\
    TI (blemished) 
    & 0.217 & 0.852 & 0.576 & 0.909 & 0.263\\
    \sy{Ours (TI-based)}
    & \textbf{0.337} & \textbf{1.300} & \textbf{0.649} & \textbf{1.020} & \textbf{0.282}\\
    \bottomrule
\end{tabular}
\caption{Quantitative results - ID.}
\label{tab:quantitative-ID}
\end{table}

\begin{table}[t]
\centering
\fontsize{9pt}{9pt}\selectfont
\setlength{\tabcolsep}{3pt}
\begin{tabular}{l @{\hskip 0.1in} ccccc  @{\hskip 0.1in} ccccc}
    \toprule
    \multirow{2}{*}{Method} &
      \multicolumn{5}{c}{\texttt{\WMModel on WM-OOD-test}} \\
      \cmidrule(lr{0.1in}){2-6}
      & {I\textsuperscript{DINO}$\uparrow$} & {R\textsuperscript{DINO}$\uparrow$} & {I\textsuperscript{CLIP}$\uparrow$} & {R\textsuperscript{CLIP}$\uparrow$} & {T\textsuperscript{CLIP}$\uparrow$} \\
      \midrule
    {\color{lightgray} TI (unblemished)} 
    & {\color{lightgray}0.488} & {\color{lightgray}1.278} & {\color{lightgray}0.730} & {\color{lightgray}1.136} & {\color{lightgray}0.283} \\
    TI (blemished)
    & 0.229 & 0.858 & 0.575 & 0.929 & 0.262 \\
    \sy{Ours (TI-based)} 
    & \textbf{0.356} & \textbf{1.237} & \textbf{0.654} & \textbf{1.079} & \textbf{0.282} \\
    \bottomrule
\end{tabular}
\caption{Quantitative results - OOD.}
\label{tab:quantitative-OOD}
\end{table}
\section{Experiment}
\label{sec:experiment}
\subsection{Implementation Details}
We employ the pre-trained LDM~\cite{rombach2022high} following the official implementation of Textual Inversion~\cite{gal2022image} as our base diffusion model. We train the blemished textual embeddings for 5k steps using Textual Inversion.
We use a learning rate of 5e-3 to optimize our Artifact-free embedding and 3e-5 for the partial fine-tuning of key and value weights. Note that all other parameters within the pre-trained diffusion model remain frozen. All experiments are conducted on 2 NVIDIA RTX 3090 GPUs. In the main paper, we focus on the comparison with Textual Inversion and DreamBooth to demonstrate the efficiency of our proposed contributions. See the Appendix for additional comparisons and applications.

\subsection{Evaluation Benchmark}
\label{subsec: benchmark}
\paragraph{\bf Test dataset} 
We construct the test dataset using 16 novel subjects that differ from the subjects in the training set. These subjects encompass a wide range of categories, including pets, plants, toys, transportation, furniture, and wearable items. We form the visible test artifacts into two categories: (1) in-distribution watermarks (\texttt{WM-ID-test}) containing the same type as the training data, and (2) out-of-distribution watermarks (\texttt{WM-OOD-test}) of different types from the training data.
Within the \texttt{WM-ID-test} and \texttt{WM-OOD-test}, we synthesize 5 distinct artifacts for each category, resulting in 80 test sets. 

\begin{figure} [t!]
  \centering  \includegraphics[width=1.0\linewidth]{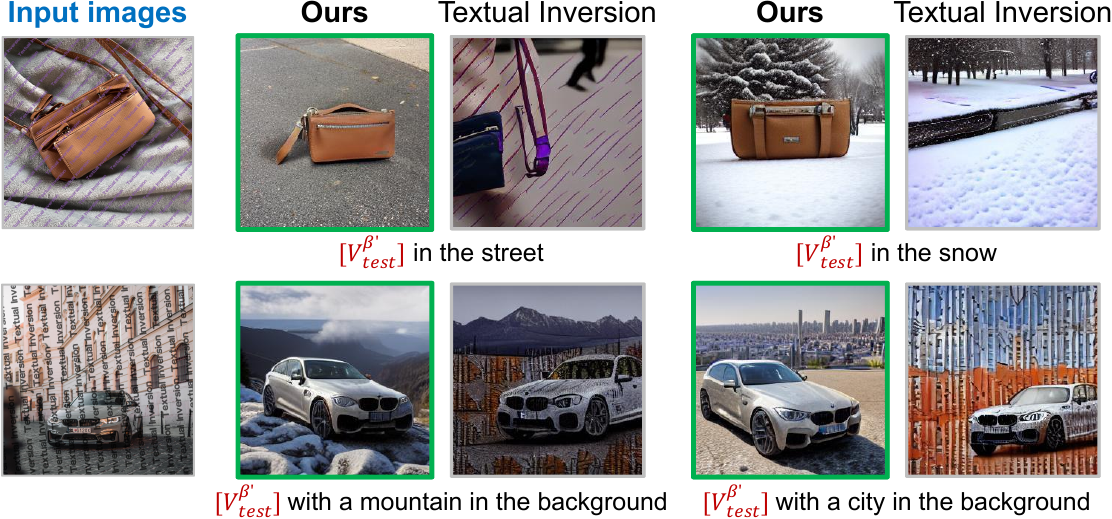}
  \caption{Qualitative Comparison - ID.
  Unlike Textual Inversion which struggles to produce reasonable generation from blemished inputs, our method (\WMModel) consistently learns the distinguished features of the given subject and achieves high-quality generation without distortion.}
  
  \label{fig:quality-ID}
\end{figure}
\begin{figure} [t!]
  \centering  \includegraphics[width=1.0\linewidth]{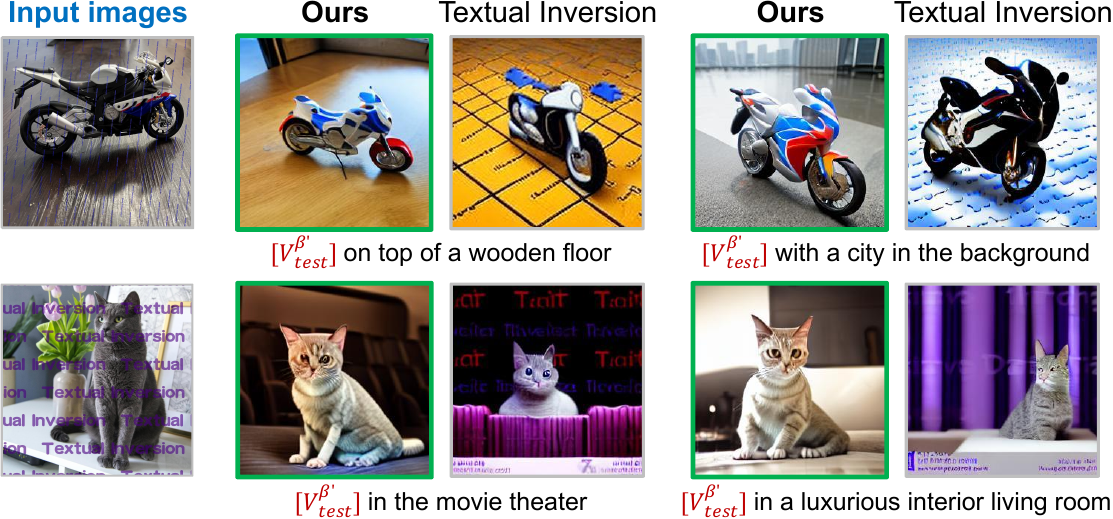}
  \caption{Qualitative Comparison - OOD. Our method (\WMModel) is generalizable to process out-of-distribution artifacts that are unseen during the \sy{fine-tuning}, demonstrating much better performance than Textual Inversion. Best viewed in PDF with zoom.}
  \label{fig:quality-OOD}
\end{figure}
\paragraph{\bf Evaluation metrics} 
We evaluate the performance of blemished subject-driven generation from three perspectives: (1) the fidelity of subject reconstruction, (2) the fidelity of text conditioning, and (3) the effectiveness of mitigating the negative impacts of artifacts. Following common practice~\cite{gal2022image,ruiz2023dreambooth}, we use CLIP~\cite{radford2021learning} and DINO~\cite{caron2021emerging} similarities for measuring these metrics. For the first metric, we calculate the CLIP and DINO similarity between the generated images and the unblemished version of the input images, respectively denoted as I\textsuperscript{CLIP} and I\textsuperscript{DINO}. For the second metric, we calculate the CLIP similarity between the generated images and the text prompt, denoted as T\textsuperscript{CLIP}. For the third metric, we calculate the relative ratio of similarities between generated images and unblemished input images compared to their blemished versions, defined as
\begin{equation}
    \text{R}^\text{CLIP} = \text{I}^\text{CLIP} / \text{I}^\text{CLIP}_\beta \quad \text{R}^\text{DINO} = \text{I}^\text{DINO} / \text{I}^\text{DINO}_\beta
    \label{eq:Ratio}
\end{equation}
where $\text{I}^\text{CLIP}_\beta$ and $\text{I}^\text{DINO}_\beta$ respectively denote CLIP and DINO similarities between the generated images and the \emph{blemished} input images. A relative ratio greater than 1 indicates that generated images resemble unblemished images more than blemished counterparts, suggesting fewer artifacts. Conversely, a ratio less than 1 indicates that generated images are heavily distorted with more artifacts. We use DINO ViT-S/16~\cite{caron2021emerging} and CLIP ViT-B/32~\cite{radford2021learning} to compute all metrics.

\subsection{Quantitative Comparisons} \label{subsec: quantitative}
We conduct both in-distribution and out-of-distribution quantitative evaluations of our method and compare it to Textual Inversion with blemished embeddings. We additionally report the results using Textual Inversion on unblemished images as a reference, although it is not a direct comparison to our model.

\paragraph{\bf In-distribution (ID) analysis} We consider the in-distribution scenarios by testing \WMModel on \texttt{WM-ID-test}.
In Tab. \ref{tab:quantitative-ID}, we can observe that the use of blemished embeddings in Textual Inversion leads to comprehensive performance decline including: (1) lower subject reconstruction fidelity (\ie, I\textsuperscript{DINO} and I\textsuperscript{CLIP}) due to the subject distortion in image generation; (2) lower efficiency for artifact removal (\ie, R\textsuperscript{DINO} and R\textsuperscript{CLIP}) due to inability to remove artifacts; (3) lower prompt fidelity (\ie, T\textsuperscript{CLIP}) since the prompt-guided background is unrecognizable due to blemishing artifacts.
In contrast, our method consistently achieves higher scores than Textual Inversion with blemished embeddings across the board, demonstrating the efficiency of \VName in various aspects.

\paragraph{\bf Out-of-distribution (OOD) analysis} We pleasantly discover that \WMModel possesses the capability to handle out-of-distribution scenarios, owing to its training with watermarks of diverse types.
We consider the out-of-distribution (OOD) scenarios for \WMModel by testing it on \texttt{WM-OOD-test},
as presented in Tab.~\ref{tab:quantitative-OOD}. Similar to ID evaluation, all of our metrics yield higher results than Textual Inversion with blemished embeddings. These results further demonstrate the generalizability of our method.

\subsection{Qualitative Comparisons} \label{subsec: qualitative} 
We present qualitative comparisons between the output generated via ArtiFade and Textual Inversion with blemished textual embeddings, including in-distribution scenarios in Fig. \ref{fig:quality-ID} and out-of-distribution scenarios in Fig. \ref{fig:quality-OOD}. 
\paragraph{\bf In-distribution analysis}
The images generated by Textual Inversion exhibit noticeable limitations when using blemished textual embeddings. Specifically, as depicted in Fig. \ref{fig:quality-ID}, all rows predominantly exhibit cases of incorrect backgrounds that are highly polluted by watermarks. By using ArtiFade, we are able to eliminate the background watermarks.

\paragraph{\bf Out-of-distribution analysis} 
In addition, we conduct experiments with our \WMModel to showcase its capability to remove out-of-distribution watermarks, as shown in Fig. \ref{fig:quality-OOD}. It is important to note that in the first row, the watermark in the input images may not be easily noticed by human eyes upon initial inspection due to the small font size and high image resolution. However, these artifacts have a significant effect when used to train blemished embeddings for generating images. ArtiFade effectively eliminates the artifacts on the generated images, improving reconstruction fidelity and background accuracy, hence leading to substantial enhancements in overall visual quality.

\subsection{ArtiFade with DreamBooth}
\begin{table}[t]
\centering
\fontsize{9pt}{9pt}\selectfont
\setlength{\tabcolsep}{4pt}
\begin{tabular}{l @{\hskip 0.2in} ccccc}
    \toprule
    \multirow{2}{*}{Method} &
      \multicolumn{5}{c}{\texttt{WM-ID-test}}  \\
      \cmidrule(lr{0.1in}){2-6}
      & {I\textsuperscript{DINO}$\uparrow$} & {R\textsuperscript{DINO}$\uparrow$} & {I\textsuperscript{CLIP}$\uparrow$} & {R\textsuperscript{CLIP}$\uparrow$} & {T\textsuperscript{CLIP}$\uparrow$} \\
      \midrule
    \color{lightgray}TI (unblemished) 
    & \color{lightgray}0.488 & \color{lightgray}1.349 & \color{lightgray}0.730 & \color{lightgray}1.070 & \color{lightgray}0.283 \\
    TI (blemished) 
    & 0.217 & 0.852 & 0.576 & 0.909 & 0.263 \\

    DB (blemished)
    & 0.503 & 0.874 & 0.738 & 0.939 & 0.272 \\

    Ours (TI-based)
    & 0.337 & 1.300 & 0.649 & 1.020 & 0.282 \\
    
    Ours (DB-based)
    & \textbf{0.589} & \textbf{1.308} & \textbf{0.795} &\textbf{1.083} & \textbf{0.284} \\
    \bottomrule
\end{tabular}
\caption{Quantitative comparison with DreamBooth.}
\label{tab:quantitative-ID-DB}
\end{table}
\begin{figure} [t]
  \centering  \includegraphics[width=1.0\linewidth]{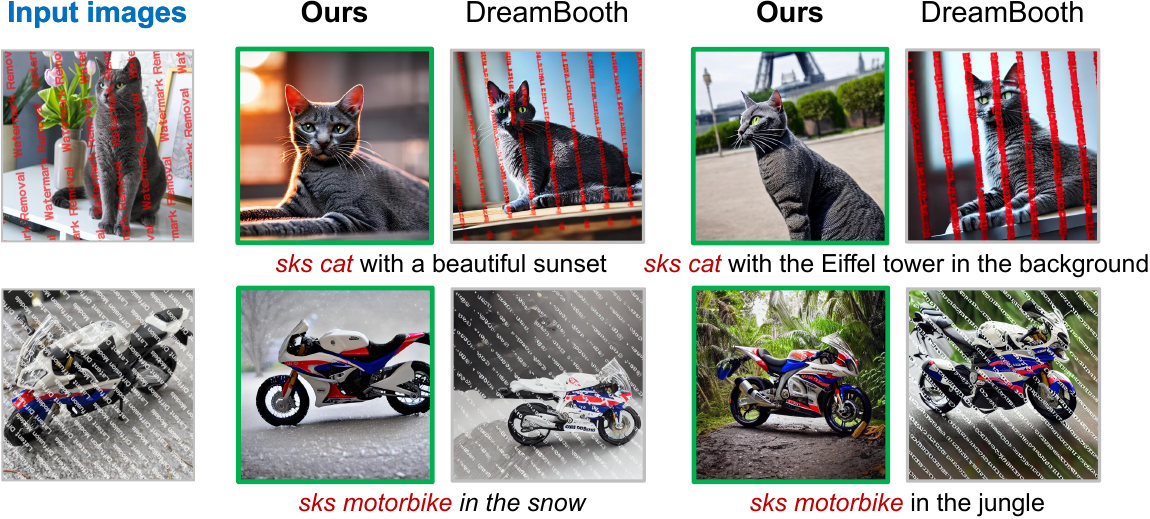}
  \caption{Qualitative comparison with DreamBooth.}
  \label{fig:quality-ID-DB}
\end{figure}
The ArtiFade fine-tuning framework is not limited to Textual Inversion with textual embedding; it can also be generalized to DreamBooth. We use the same training dataset and blemished subsets as in the case of the \WMModel (\ie, $N=$ 20, $L_{WM}$= 10). The vanilla DreamBooth fine-tunes the whole UNet model, which conflicts with the fine-tuning parameters of ArtiFade. We therefore use DreamBooth with low-rank approximation (LoRA)\footnote{\url{https://huggingface.co/docs/peft/main/en/task_guides/dreambooth_lora}} to train LoRA adapters~\cite{hu2021lora} for the text encoder, value, and query weights of the diffusion model for each blemished subset using Stable Diffusion v1-5. For simplicity, we will use DreamBooth to refer to DreamBooth with LoRA below. During the fine-tuning of DreamBooth-based ArtiFade, we load the pre-trained adapters and only unfreeze key weights since value weights are reserved for DreamBooth subject information. In Tab. \ref{tab:quantitative-ID-DB}, it is evident that our method, based on DreamBooth, yields the highest scores among all cases. Our method also maintains DreamBooth's advantages in generating images with higher subject fidelity and more accurate text prompting, outperforming ArtiFade with Textual Inversion. We show some qualitative results in Fig. \ref{fig:quality-ID-DB}.

\subsection{Invisible Artifacts Blemished Subject Generation}
\begin{figure} [t]
  \centering  \includegraphics[width=1.0\linewidth]{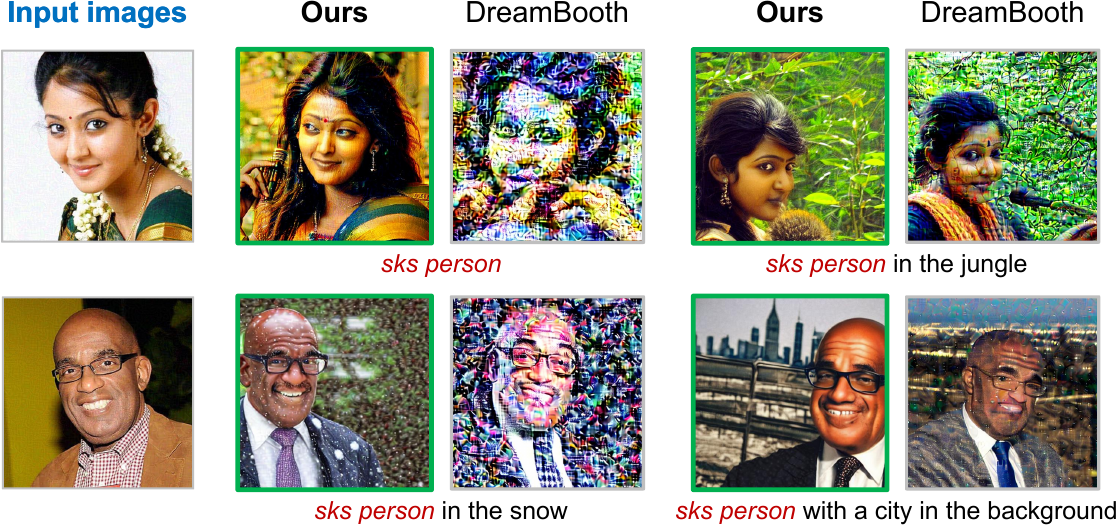}
  \caption{Qualitative Comparison between ours and DreamBooth when inputs are blemished by invisible adversarial noises.}
  \label{fig:quality-ID-DB-adversarial}
\end{figure}

ArtiFade demonstrates exceptional performance in handling subjects characterized by intricate features and blemished by imperceptible artifacts. We collect 20 human figure datasets from the VGGFace2 dataset~\cite{cao2018vggface2}. We then use the Anti-DreamBooth~\cite{van2023anti} ASPL method to add adversarial noises to each group of images, producing 20 blemished datasets for fine-tuning a DreamBooth-based ArtiFade model. The model is fine-tuned for 12k steps. As illustrated in Fig.~\ref{fig:quality-ID-DB-adversarial}, our approach surpasses the DreamBooth in differentiating the learning of adversarial noises from human face features. In contrast to DreamBooth, which is fooled into overfitting adversarial noises, thereby generating images with a heavily polluted background, our model reconstructs human figures in image generation while maintaining high fidelity through text prompting.

\subsection{Ablation Studies} 
\label{subsec: AblationStudy}
\begin{figure}[t]
    \centering
        \includegraphics[width=1.0\linewidth]{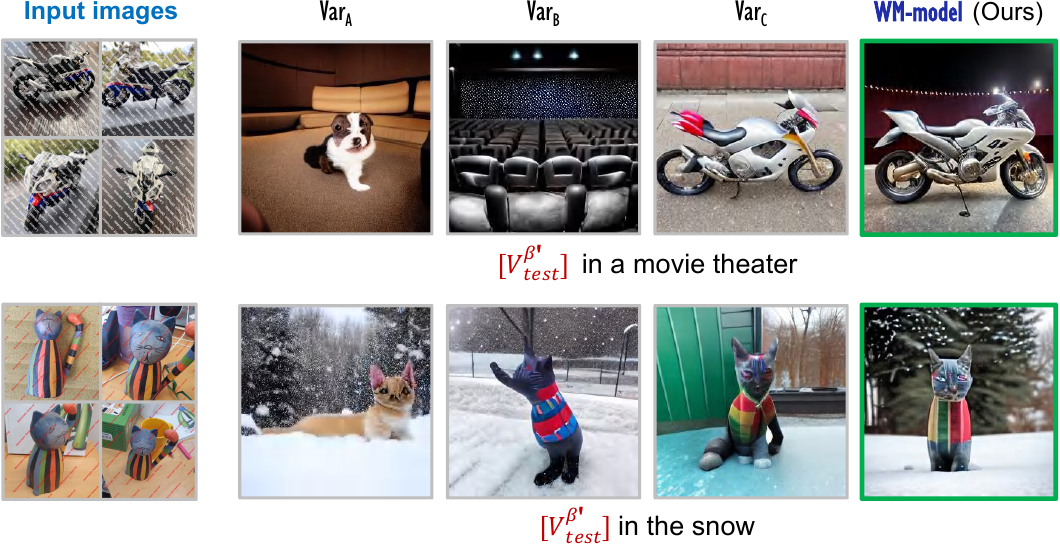}
        \caption{Qualitative comparison of ablation studies.
        }
        \label{fig:ablation}
    \centering
\end{figure}

We conduct ablation studies to demonstrate the efficiency of our method by comparing with three alternative variants, which encompass (1) $\mathtt{Var_A}$, where we solely fine-tune the artifact-free embedding; (2) $\mathtt{Var_B}$, where we fine-tune parameters related to image features, \ie, query weights $W^q$, along with the artifact-free embedding, and (3) $\mathtt{Var_C}$, where we fine-tune key and value weights, \ie, $W^k$ and $W^v$, exclusively. We use our $\WMModel$ to compare it with other variants by testing on the \texttt{WM-ID-test}.
\paragraph{Effect of partial fine-tuning} As shown in Tab. \ref{tab:ablation}, compared to $\mathtt{Var_A}$, our full method yields higher scores on all metrics by a significant margin, except for R\textsuperscript{DINO}. This is reasonable, as the artifact-free embedding can be easily overfitted to the training data, resulting in generated images that resemble a fusion of training images (Fig. \ref{fig:ablation}, $\mathtt{Var_A}$). As a result, the denominator of R\textsuperscript{DINO}, namely the similarity between the generated image and the blemished image, is significantly decreased, leading to a high R\textsuperscript{DINO}. Due to similar reason, $\mathtt{Var_A}$ shows lowest I\textsuperscript{DINO}, I\textsuperscript{CLIP}, and T\textsuperscript{CLIP} among all variants, indicating that it fails to reconstruct the correct subject.  
Overall, both quantitative and qualitative evaluation showcases that solely optimizing the artifact-free embedding is insufficient to capture the distinct characteristics presented in the blemished input image, demonstrating the necessity of partial fine-tuning.

\paragraph{\bf Effect of fine-tuning key and value weights} As shown in Tab. \ref{tab:ablation} and Fig. \ref{fig:ablation}, $\mathtt{Var_B}$ yields unsatisfactory outcomes in all aspects compared to ours. The lower R\textsuperscript{DINO} and R\textsuperscript{CLIP} suggest that the generated images retain artifact-like features and bear closer resemblances to the blemished subsets. Furthermore, the reduced T\textsuperscript{CLIP} indicates diminished prompt fidelity, as the approach fails to accurately reconstruct the subject from the blemished embeddings, which is also evidenced by Fig.~\ref{fig:ablation}. These findings suggest that fine-tuning the parameters associated with text features yields superior enhancements in terms of artifact removal and prompt fidelity.
\begin{table}[t]
\fontsize{9pt}{9pt}\selectfont
  \centering
\setlength{\tabcolsep}{3pt}
  \begin{tabular}{l|ccc|ccccc}
    \toprule
    Method & $W^{kv}$ & $W^q$ & \VEmbed & $\text{I}^\text{DINO}$ & $\text{R}^\text{DINO}$ & $\text{I}^\text{CLIP}$ & $\text{R}^\text{CLIP}$ & $\text{T}^\text{CLIP}$\\
    \midrule
    $\mathtt{Var_A}$   & & & \checkmark 
        & 0.154 & \textbf{1.412} & 0.566 & 0.984 & 0.265\\
    $\mathtt{Var_B}$    & & \checkmark & \checkmark    
         & 0.283 & 1.230 & 0.617 & 0.978 & 0.277\\
    $\mathtt{Var_C}$   & \checkmark & &  
        & \textbf{0.342} & 1.292 & \textbf{0.652} & 1.019 & 0.280\\
   
    Ours   & \checkmark & & \checkmark    
        & 0.337 & 1.300 & 0.649 & \textbf{1.020} & \textbf{0.282}\\
    \bottomrule
  \end{tabular}
    \caption{Quantitative comparison of ablation studies.}
  \label{tab:ablation}
\end{table}
\paragraph{\bf Effect of the artifact-free embedding} With $\mathtt{Var_C}$, we exclude the optimization of artifact-free embedding. In Tab.~\ref{tab:ablation}, we can observe that $\mathtt{Var_C}$ yields higher I\textsuperscript{DINO} and I\textsuperscript{CLIP} but lower R\textsuperscript{DINO} and R\textsuperscript{CLIP} compared to our \WMModel, which indicates that the approach achieves higher subject fidelity but lower efficiency in eliminating artifacts when generating images. Since our primary objective is to generate artifact-free images from blemished textual embedding, our \WMModel chooses to trade off subject reconstruction fidelity for the ability to remove artifacts. Additionally, this approach produces lower T\textsuperscript{CLIP} than ours, suggesting that the artifact-free embedding effectively improves the model's capability to better preserve text information (see Fig.~\ref{fig:ablation}).
\begin{figure}[!t]
  \centering
  \begin{subfigure}{1.0\linewidth}
    \includegraphics[width=\linewidth, keepaspectratio]{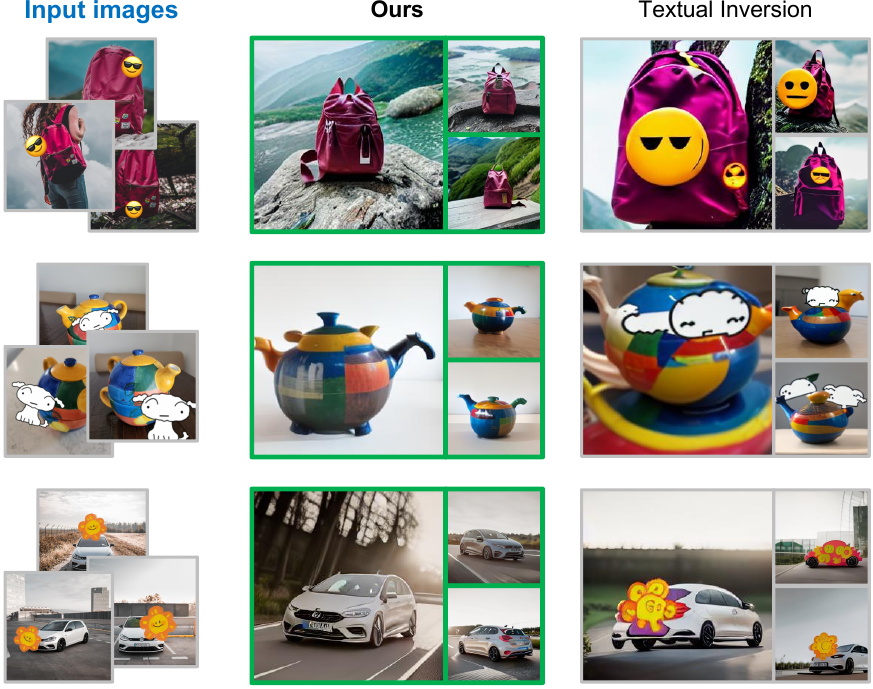}
    \caption{Sticker removal.\vspace{10pt}}
    \label{fig:sticker application}
  \end{subfigure} 
  \begin{subfigure}{1.0\linewidth}
    \includegraphics[width=\linewidth, keepaspectratio]{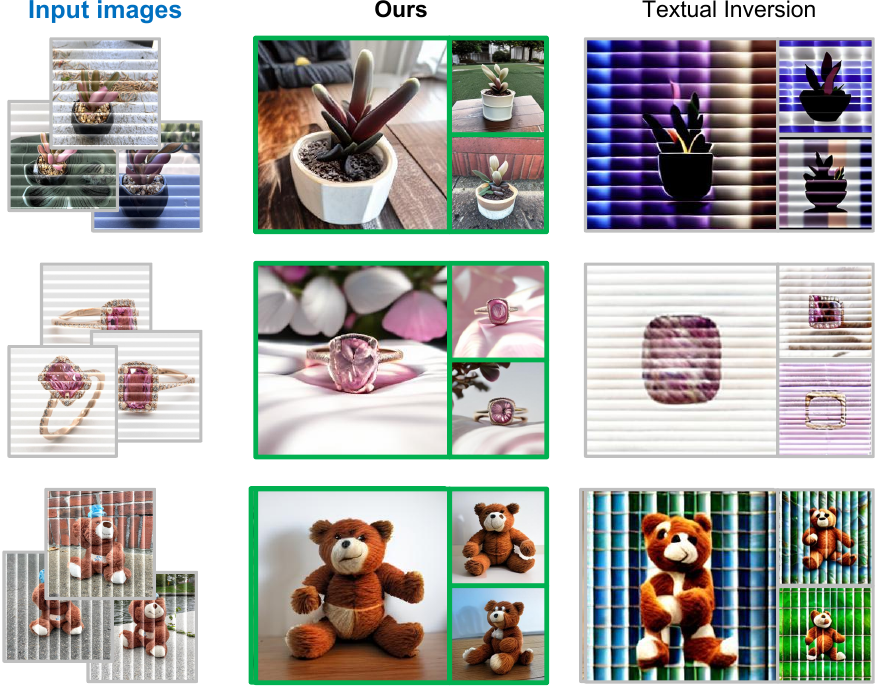}
    \caption{Glass effect removal.}
    \label{fig:glass application}
  \end{subfigure}
  \caption{Applications. Our \WMModel can be applied to remove various unwanted artifacts in the input images, \eg stickers, glass effect, etc.
  }
  \label{fig:application}
\end{figure}

\section{More Applications}
We apply our \WMModel to more artifact cases, such as stickers and glass effects, showcasing its broad applicability.
\paragraph{\bf Sticker removal.} In Fig.~\ref{fig:sticker application}, we test \WMModel on input images that are blemished by cartoon stickers.
The cartoon sticker exhibits randomized dimensions and is positioned arbitrarily within each image.
\WMModel can effectively eliminate any stickers while concurrently addressing improper stylistic issues encountered during image generation.
\paragraph{\bf Glass effect removal.} 
We further test \WMModel on input images that are blemished by glass effect in Fig.~\ref{fig:glass application}.
We apply a fluted glass effect to images to replicate real-life scenarios where individuals capture photographs of subjects positioned behind fluted glass. This glass can have specific reflections and blurring, which may compromise the overall quality of image generation when using Textual Inversion. The use of our model can fix the distortions of the subjects and the unexpected background problem, significantly improving image quality.
\section{Conclusion}
\label{sec:conclusion}
In conclusion, we introduce \VName{} to address the novel problem of generating high-quality and artifact-free images in the blemished subject-driven generation. Our approach involves fine-tuning a diffusion model along with artifact-free embedding to learn the alignment between unblemished images and blemished information.
We present an evaluation benchmark to thoroughly assess a model's capability in the task of blemished subject-driven generation. We demonstrate the effectiveness of ArtiFade in removing artifacts and addressing distortions in subject reconstruction under both in-distribution and out-of-distribution scenarios. 

\clearpage
\bibliography{main}
\clearpage
\appendix
\renewcommand\thesection{\Alph{section}}
\renewcommand\thesubsection{\thesection.\Alph{subsection}}
\title{Appendix}
\maketitle
\section{Training Dataset Details}
\begin{figure}[h]
  \centering
    \includegraphics[width=\linewidth]{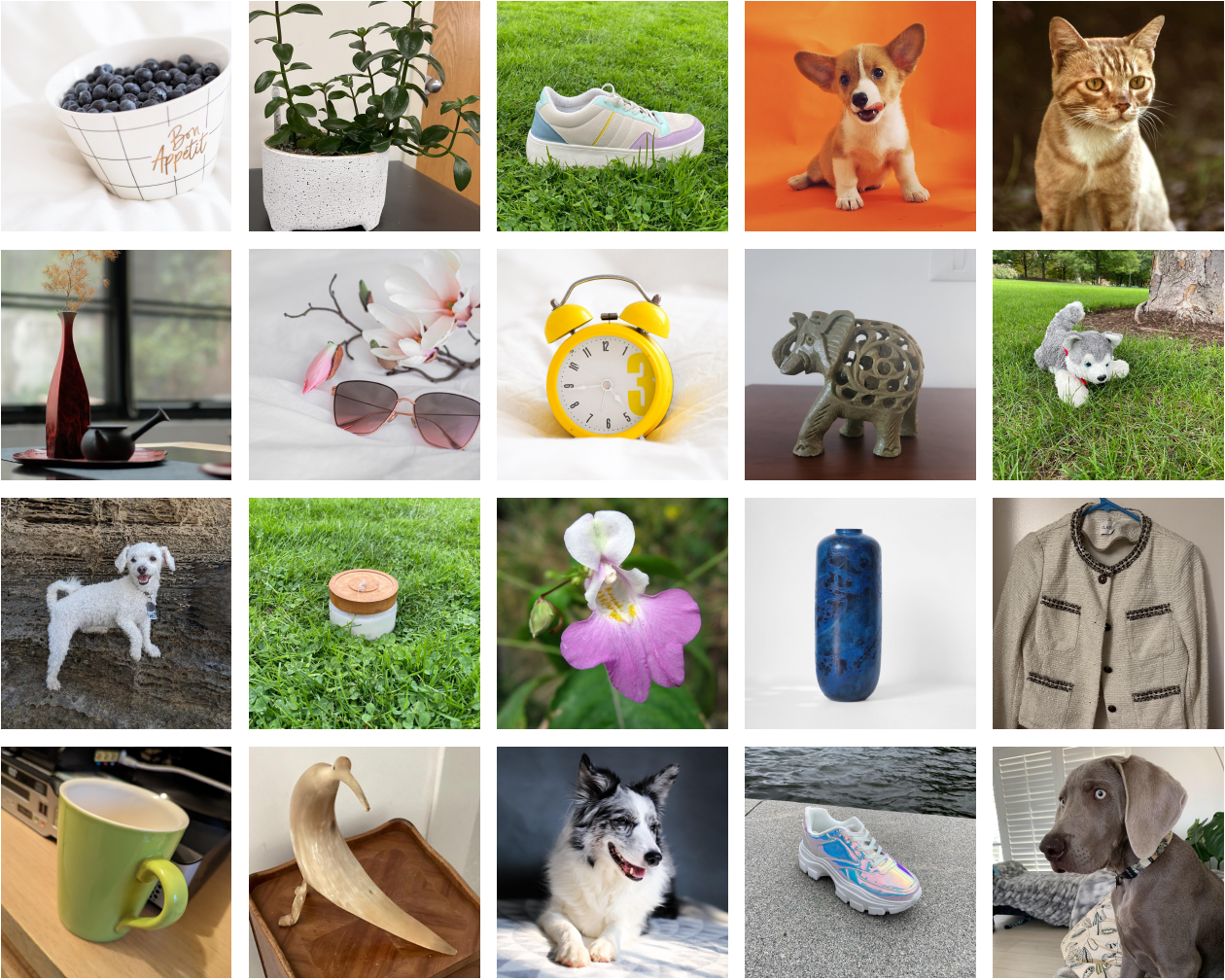}
    \caption{Examples of unblemished training images. We show a total of 20 images, each containing one distinct subject.\vspace{-20pt}}
    \label{fig:subjects-example}
\end{figure}

\begin{figure*}[!ht]
  \centering
    \includegraphics[width=0.9\linewidth, keepaspectratio]{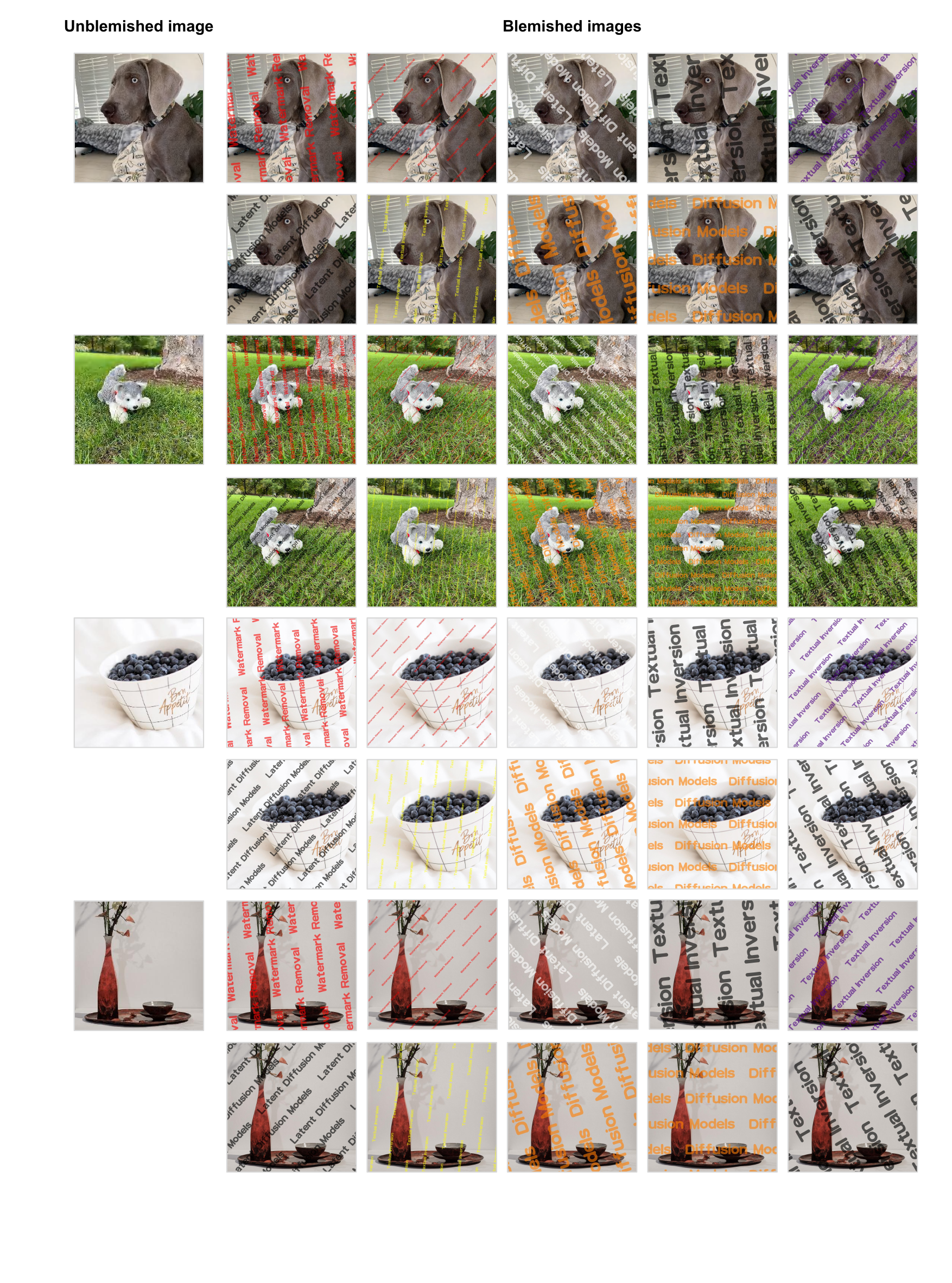}
    \caption{Examples of the training dataset: unblemished images and their corresponding blemished images.}
    \label{fig:unblemished-and-blemished}
\end{figure*}

Our training dataset consists of 20 training subjects, used for the fine-tuning stage of our \VName models. We show an example image of each subject in Fig.~\ref{fig:subjects-example}. In Fig.~\ref{fig:unblemished-and-blemished}, we showcase several unblemished images alongside their corresponding blemished versions, each featuring one of the 10 watermark types.

\section{Test Dataset Details}
\begin{figure}[t]
  \centering
    \includegraphics[width=\linewidth, keepaspectratio]{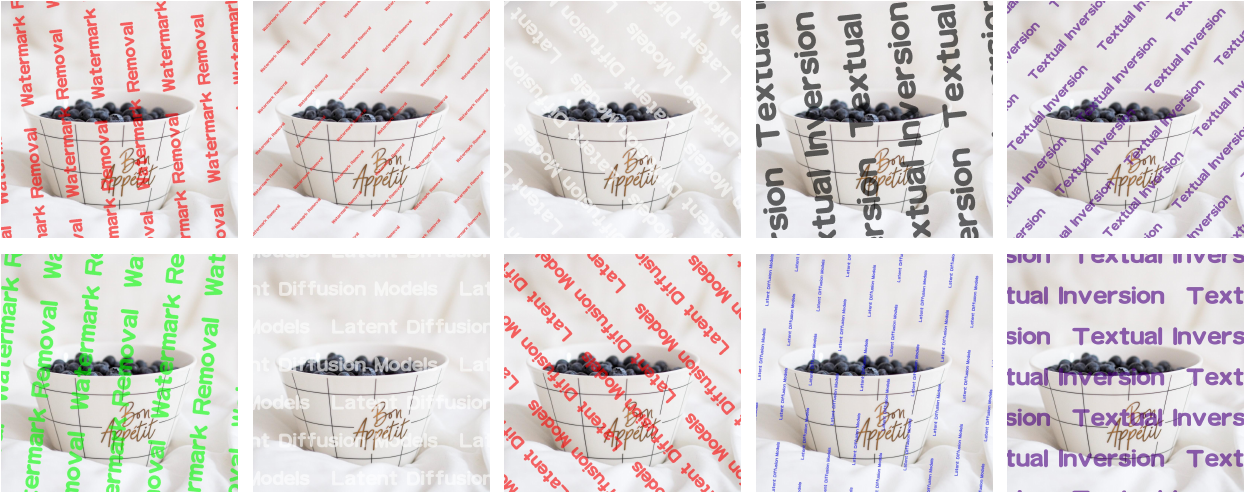}
    \caption{Example of test watermark types. The first row displays the \texttt{WM-ID-TEST}, while the second row presents the \texttt{WM-OOD-TEST}.}
    \label{fig:ID-and-OOD}
\end{figure}

In Fig.~\ref{fig:ID-and-OOD}, we illustrate our \texttt{WM-ID-TEST} watermark types (see the first row) and \texttt{WM-OOD-TEST} watermark types (see the second row). The \texttt{WM-ID-TEST} watermarks are chosen from the training watermarks displayed in Fig.~\ref{fig:unblemished-and-blemished}. On the other hand, the \texttt{WM-OOD-TEST} watermarks differ in font size, orientation, content, or color from all the training watermarks presented in Fig.~\ref{fig:unblemished-and-blemished}.

\section{Analysis of Watermark Density}
\begin{figure}[!ht]
    \centering
    \includegraphics[width=0.9\linewidth]{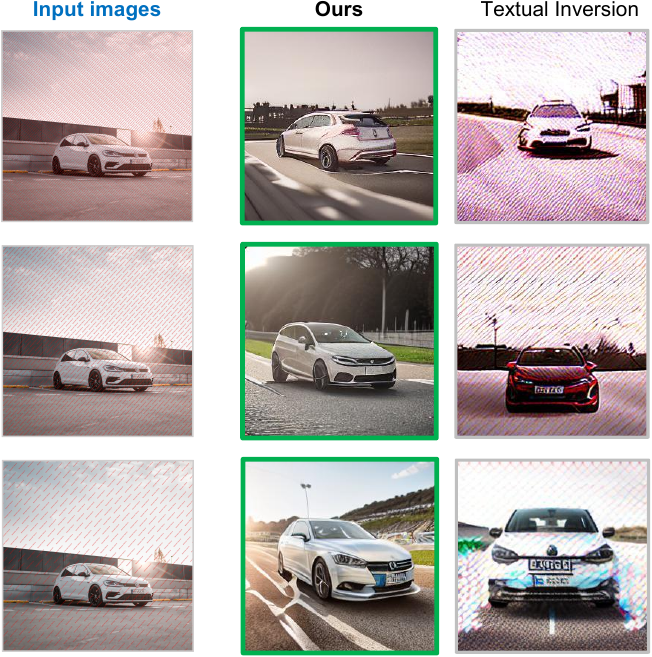} 
   \caption{Varying qualities of input images. Our method (\WMModel) can be used to remove watermarks when input images are of any quality.}
   \label{fig:density}
\end{figure}

In Fig.~\ref{fig:density}, we present results to illustrate the impact of varying watermark densities (\ie, varying qualities), highlighting the robust ability of our \WMModel to remove watermarks under all conditions.
\begin{figure}[!ht]
    \centering
        \includegraphics[width=1.0\linewidth]{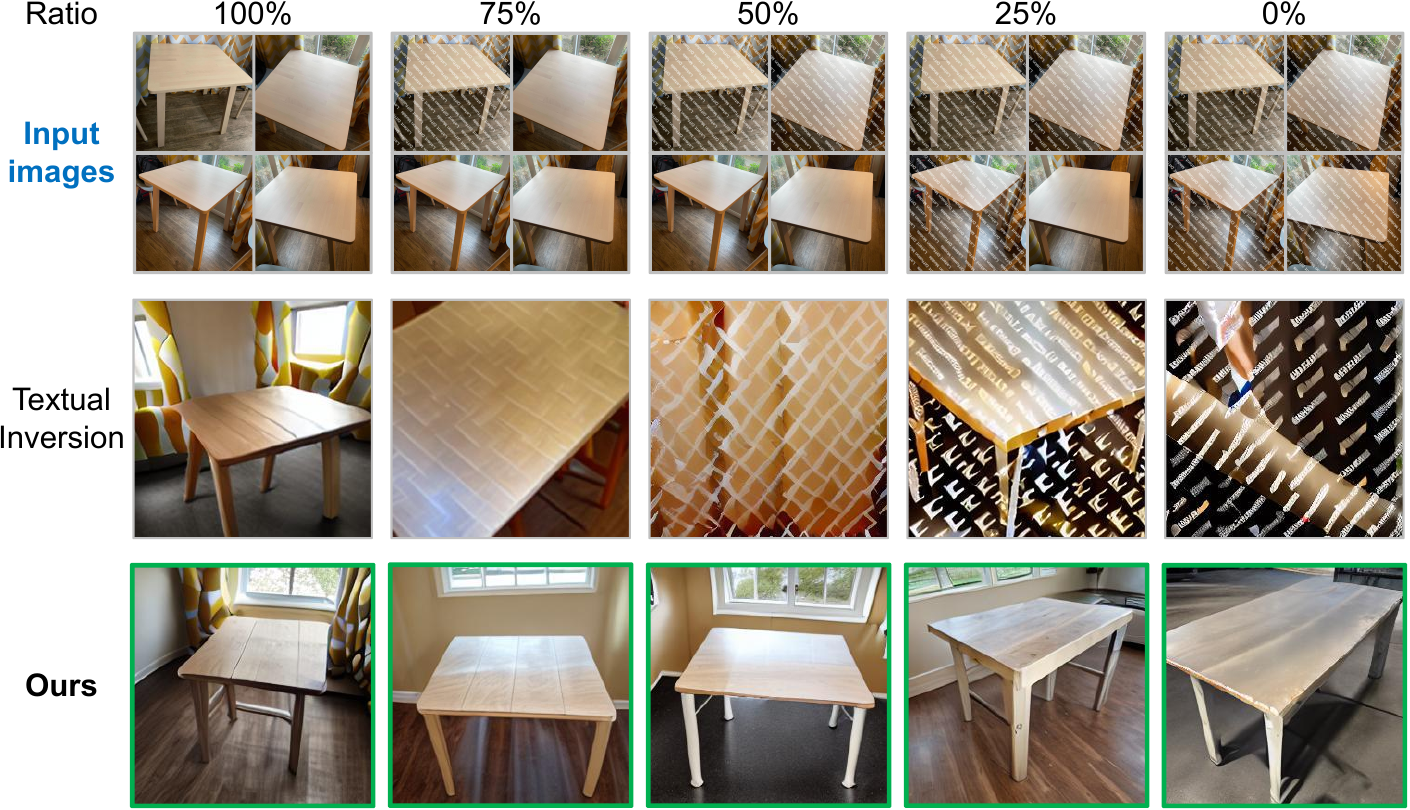}
        \caption{Comparison between different ratios of unblemished images. \VName can perform well under any scenarios with different ratios of unblemished images.
        }
        \label{fig:ratio}
    \centering
\end{figure}

\section{Analysis of Unblemished Image Ratio}
We employ our \WMModel to evaluate the performance when the input images contain different proportions of unblemished images. We test our \WMModel and Textual Inversion on five ratios of unblemished images: 100\%, 75\%, 50\%, 25\%, and 0\%. The results are shown in Fig.~\ref{fig:ratio}. 

Notably, even when there is only one blemished image in the second column example, the impact on Textual Inversion is already evident, which deteriorates as the ratio decreases. Instead, our method effectively eliminates artifacts in all settings of unblemished image ratio, demonstrating its versatility in real-life scenarios.


\section{Analysis of Training Dataset Size}
We conduct an analysis to investigate the impact of the number of training subjects (\ie, the size of the training dataset) on the performance of our model. We utilize the same set of artifacts $L_{\text{WM}} = 10$, as described in Method in the main paper. We construct blemished training datasets in four different sizes: (1) with 5 subjects, (2) with 10 subjects, (3) with 15 subjects, and (4) with 20 subjects. We generate 50, 100, 150, and 200 blemished datasets for each of these cases. Subsequently, we fine-tune four distinct ArtiFade models, each with 16k training steps.

\begin{figure*}[ht]
  \centering
  \begin{subfigure}{0.45\textwidth}
    \includegraphics[width=\textwidth, keepaspectratio]{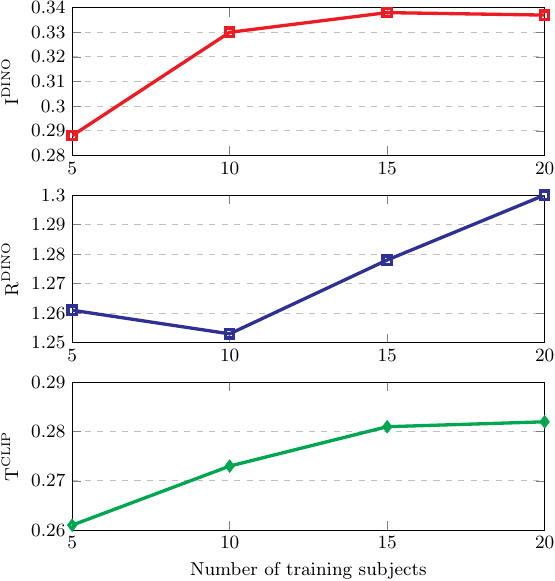}
    \caption{ID}
    \label{fig:quantitative-ID-SZ}
  \end{subfigure}\ \ 
  \begin{subfigure}{0.45\textwidth}
    \includegraphics[width=\textwidth, keepaspectratio]{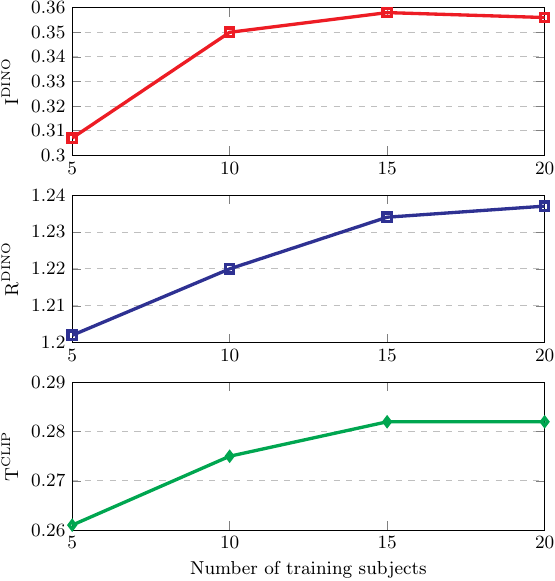}
    \caption{OOD}
    \label{fig:quantitative-OOD-SZ}
  \end{subfigure} 
  \caption{Analysis of the number of training subjects.}
\end{figure*}

\begin{figure}[!ht]
    \centering    \includegraphics[width=1.0\linewidth, keepaspectratio]{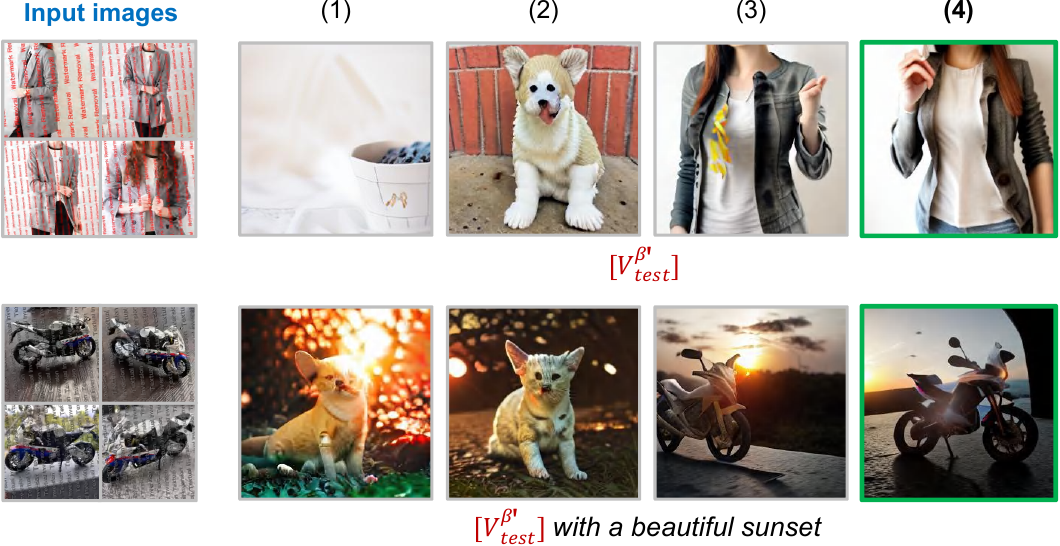}
    \caption{Qualitative results of different number of training subjects - ID.}
    \label{fig:ID-SZ}
\end{figure}

\begin{figure}[!ht]
    \centering
    \includegraphics[width=1.0\linewidth, keepaspectratio]{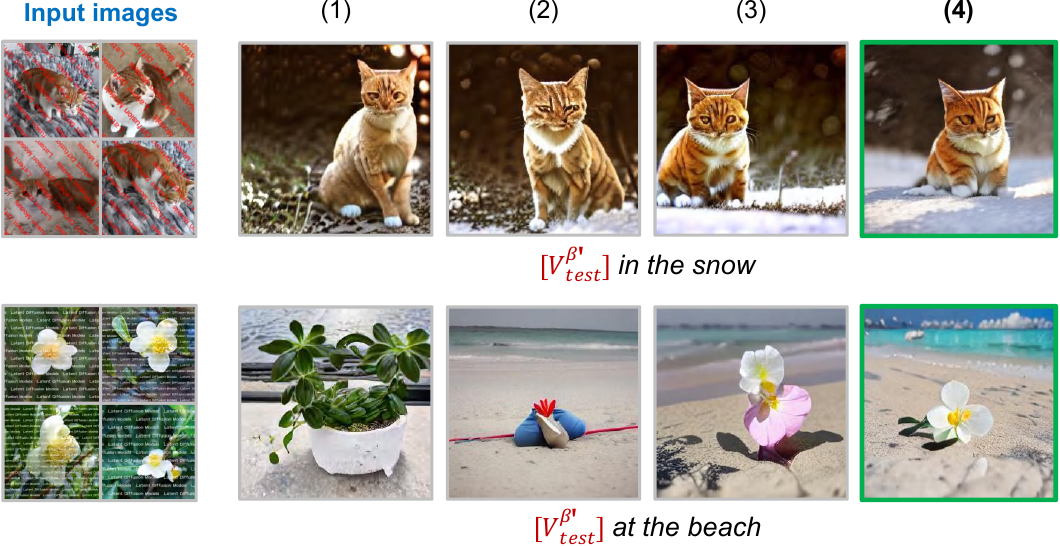}
    \caption{Qualitative results of different number of training subjects - OOD.\vspace{-10pt}}
    \label{fig:OOD-SZ}
\end{figure}

We compare the models trained using different data sizes under the in-distribution scenario (see Fig.~\ref{fig:quantitative-ID-SZ}) and under the out-of-distribution scenario (see Fig.~\ref{fig:quantitative-OOD-SZ}). We note that when the number of training subjects is less than 15, I\textsuperscript{DINO} and T\textsuperscript{CLIP} are relatively lower than the other two cases in both ID and OOD scenarios. This observation can be attributed to a significant likelihood of subject or background overfitting during the reconstruction and image synthesis processes, as visually illustrated in Fig.~\ref{fig:ID-SZ} and Fig.~\ref{fig:OOD-SZ}. However, as the number of training subjects reaches or exceeds 15, we observe a convergence in the values of I\textsuperscript{DINO} and T\textsuperscript{CLIP}, indicating a reduction in subject overfitting. Regarding R\textsuperscript{DINO}, we note that all cases exhibit values greater than one, with a slightly increasing trend as the number of training subjects rises. 

\section{Failure Cases}
\begin{figure*}[t]
  \centering
  \begin{subfigure}{0.75\textwidth}
    \includegraphics[width=\textwidth, keepaspectratio]{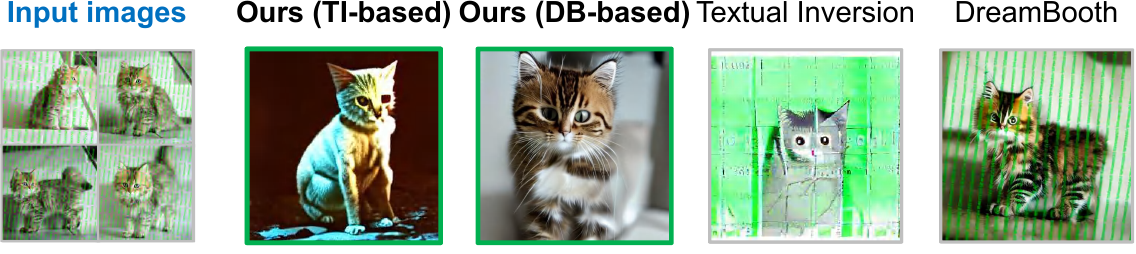}
    \caption{Incorrect subject color}
    \label{fig:incorrect_subject_color}
  \end{subfigure}
  \begin{subfigure}{0.75\linewidth}
    \includegraphics[width=\linewidth, keepaspectratio]{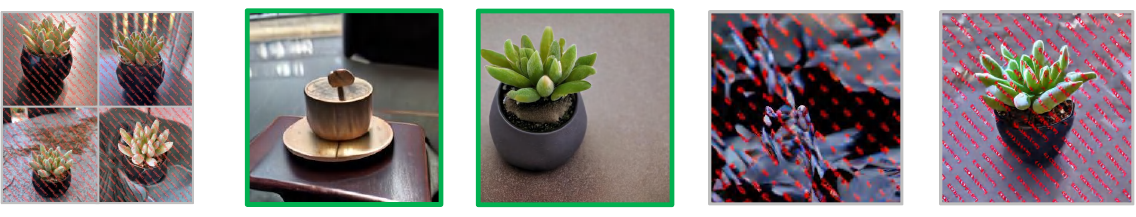}
    \caption{Incorrect subject identity}
    \label{fig:incorrect_subject}
  \end{subfigure} 
  \begin{subfigure}{0.75\linewidth}
    \includegraphics[width=\linewidth, keepaspectratio]{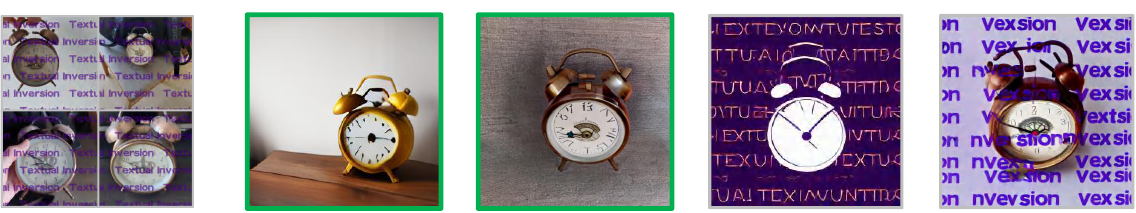}
    \caption{Subject overfitting}
    \label{fig:subject_overfitting}
  \end{subfigure} 
  \caption{Failure cases of ArtiFade based on Textual Inversion. We observe three main types of failure cases of our \WMModel: (a) incorrect subject color, (b) incorrect subject identity, and (c) subject overfitting. However, those limitations can be resolved by using ArtiFade with DreamBooth-based fine-tuning.}
  \label{fig:limitation}
\end{figure*}

We present several failure cases when applying ArtiFade based on Textual Inversion. We demonstrate the limitations of our \WMModel in Fig.~\ref{fig:limitation}. Despite the model's ability to eliminate watermarks, we still encounter issues with incorrect subject color, as shown in Fig.~\ref{fig:incorrect_subject_color}, which arises due to the influence of the watermark color. We also encounter incorrect subject identity in some cases, as demonstrated in Fig.~\ref{fig:incorrect_subject}. One possible reason is that the watermarks significantly contaminate the images, causing the learning process of embedding to focus on the contaminated visual appearance instead of the intact subject. 
Another failure case is subject overfitting, as shown in Fig.~\ref{fig:subject_overfitting}. In this case, the constructed subject overfits with a similar subject type that appears in the training dataset. This problem occurs because the blemished embedding of the testing subject closely resembles some blemished embeddings of the training subjects. Surprisingly, we find those problems can be solved by using ArtiFade based on DreamBooth, which is mentioned in Sec.~4.5. Therefore, we recommend using ArtiFade based on DreamBooth when encountering the limitations mentioned above.

\section{Additional Comparison with Textual Inversion}
\begin{figure*}[t]
  \centering
  \begin{subfigure}{0.75\linewidth}
    \includegraphics[width=\linewidth, keepaspectratio]{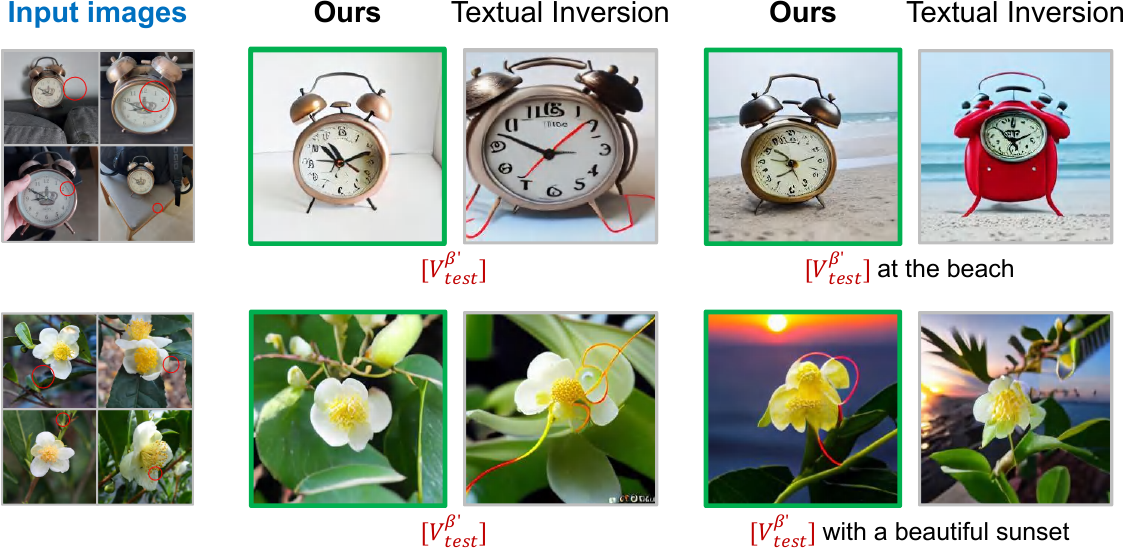}
    \caption{\RCModel on \texttt{RC-test}.\vspace{10pt}}
    \label{fig:RCModel_RCtest}
  \end{subfigure} 
  \begin{subfigure}{0.75\linewidth}
    \includegraphics[width=\linewidth, keepaspectratio]{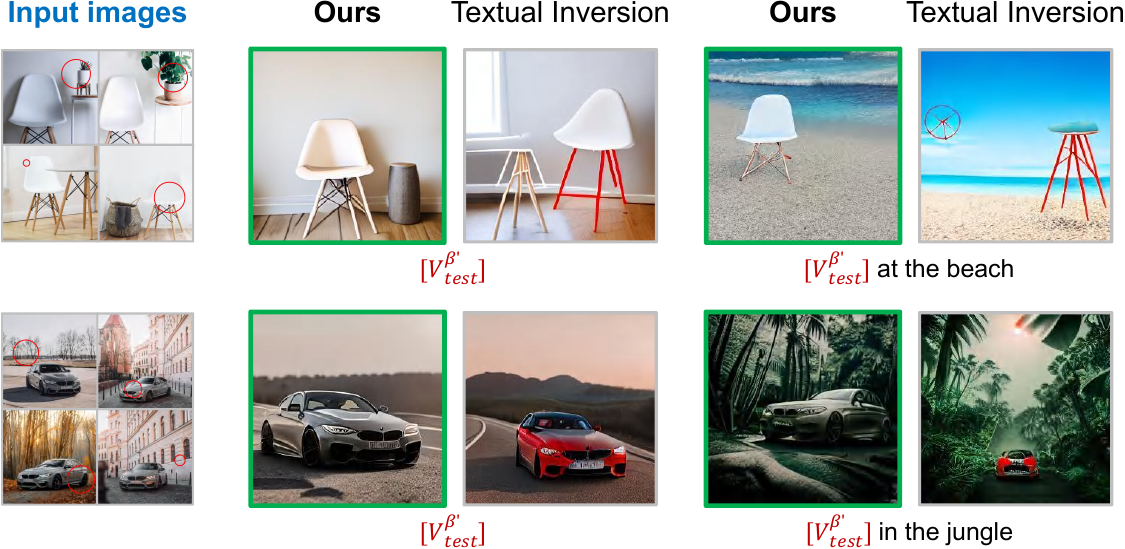}
    \caption{\WMModel on \texttt{RC-test}.}
    \label{fig:WMModel_RCtest}
  \end{subfigure}
  \caption{Qualitative results of \texttt{RC-test}. Our models consistently output high-quality and artifact-free images compared to Textual Inversion.
  }
  \label{fig:RC-qualitative}
\end{figure*}

We use the same training subjects with N$=20$ from Sec.~3.3 to train an ArtiFade model named \RCModel using red circle artifacts. For the training set of \RCModel, due to the simplicity of red circles, we only synthesize a single blemished subset (\ie, $L_{\text{RC}}=$ 1) for each subject, deriving 20 blemished subsets in total. We augment each image with a red circle mark that is randomly scaled and positioned on the source image. Considering the small scale of \RCModel's datasets, we only fine-tune \RCModel for 8k steps. 
We further introduce \texttt{RC-test}, which applies only one type of artifact (\ie, red circle) to our 16 test subjects, resulting in 16 test sets. We test both \RCModel and \WMModel on \texttt{RC-test}. The quantitative and qualitative results are shown in Tab.~\ref{tab:RC-quantitative} and Fig.~\ref{fig:RC-qualitative}, respectively.

\paragraph{\textbf{Quantitative results analysis.}}
From Tab.~\ref{tab:RC-quantitative}, we can observe that both \RCModel and \WMModel yield higher results in nearly all cases than Textual Inversion~\cite{gal2022image} with blemished inputs, showing the capability of our models to eliminate artifacts and generate subjects with higher fidelity. It is important to note that the \texttt{RC-test} is considered out-of-distribution with respect to \WMModel. Nevertheless, the metrics produced by \WMModel remain comparable to those of \RCModel, with a minor difference observed. These results provide additional evidence supporting the generalizability of our \WMModel.

\paragraph{\textbf{Qualitative results analysis.}} As illustrated in Fig.~\ref{fig:RC-qualitative}, Textual Inversion struggles with accurate color reconstruction. It also showcases subject distortions and introduces red-circle-like artifacts during image generation when using blemished embeddings. In contrast, our \RCModel (see Fig.~\ref{fig:RCModel_RCtest}) and \WMModel (see Fig.~\ref{fig:WMModel_RCtest}) are capable of generating high-quality images that accurately reconstruct the color and identities of subjects without any interference from artifacts during the image synthesis.

\begin{table}[t]
\centering
\setlength{\tabcolsep}{3pt}
\fontsize{9pt}{9pt}\selectfont
\begin{tabular}{l @{\hskip 0.2in} ccccc  @{\hskip 0.2in} ccccc}
    \toprule
    \multirow{2}{*}{Method} &
      \multicolumn{5}{c}{\texttt{RC-test}} \\
      \cmidrule(lr{0.1in}){2-6}   
      & {I\textsuperscript{DINO}} & {R\textsuperscript{DINO}} & {I\textsuperscript{CLIP}} & {R\textsuperscript{CLIP}} & {T\textsuperscript{CLIP}} \\
      \midrule
    {\color{lightgray} TI (unblemished)} 
    & {\color{lightgray}0.488} & {\color{lightgray}1.021} & {\color{lightgray}0.730} & {\color{lightgray}1.077} & {\color{lightgray}0.283}\\
    TI (blemished)
    & 0.406 & 0.990 & 0.672 & 1.042 & 0.284\\
    Ours (\RCModel)
    & \textbf{0.476} & \textbf{1.013} & 0.722 & \textbf{1.065} & \textbf{0.285}\\
    Ours (\WMModel)
    & 0.474 &  1.006 &\textbf{ 0.727} & 1.063 & 0.282\\
    \bottomrule
\end{tabular}
\caption{Quantitative results of \texttt{RC-test}.}
\label{tab:RC-quantitative}
\end{table}

\section{Additional Qualitative Comparisons}
\begin{figure*}[t]
  \centering
\includegraphics[width=0.85\textwidth, keepaspectratio]{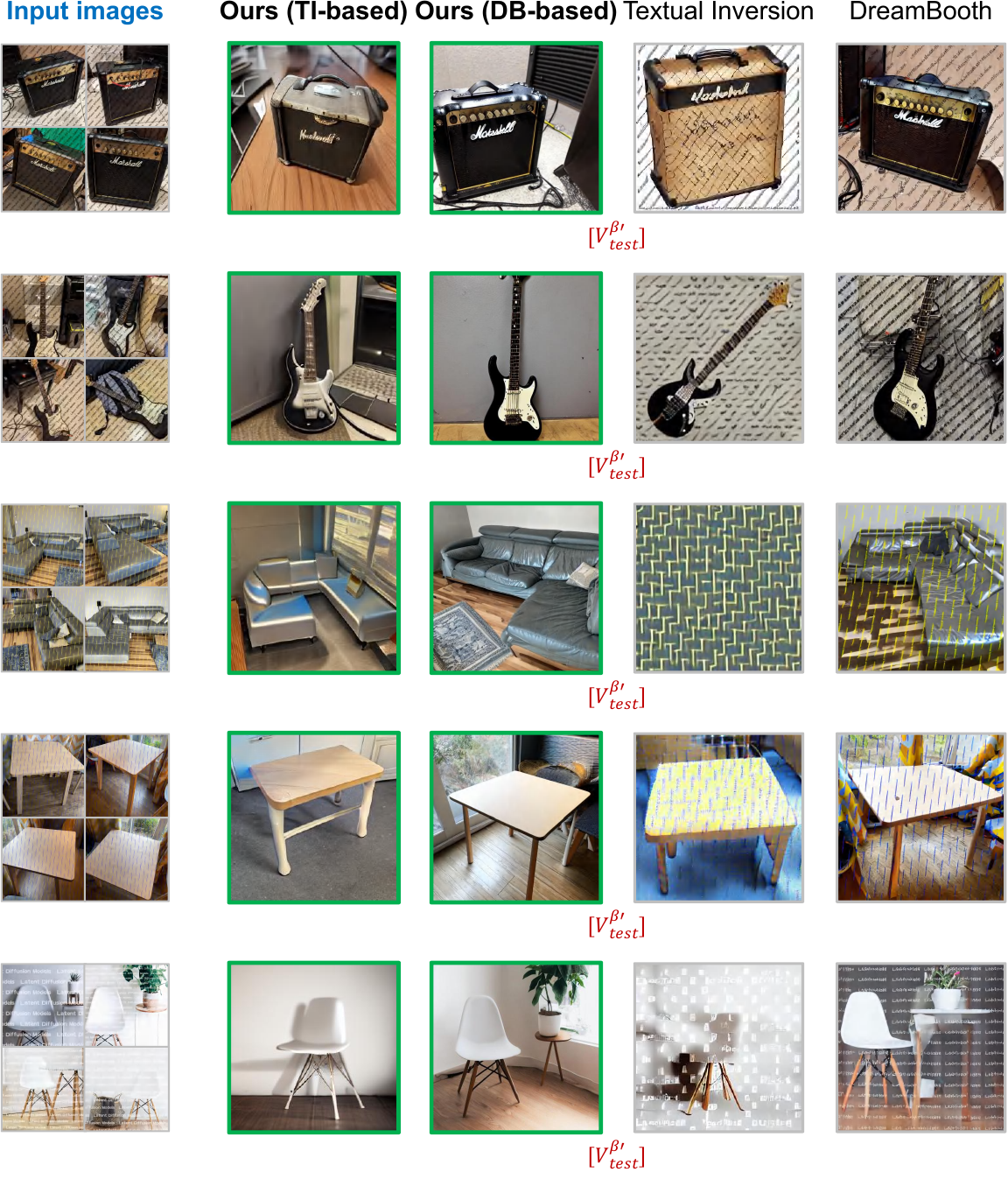}
    \caption{Additional qualitative comparisons.}
    \label{fig:More-qualitative}
\end{figure*}

We present additional qualitative results comparing our ArtiFade models with Textual Inversion~\cite{gal2022image} and DreamBooth~\cite{ruiz2023dreambooth} in Fig.~\ref{fig:More-qualitative}. We employ \WMModel and ArtiFade based on DreamBooth mentioned in Sec. 4.5.
Textual Inversion generates images with distorted subjects and backgrounds contaminated by watermarks, whereas DreamBooth can effectively capture intricate subject details and accurately reproduce watermark patterns. In contrast, our models (\ie, TI-based and DB-based ArtiFade) generate images devoid of watermark pollution with correct subject identities for both in-distribution (see the first three rows in Fig.~\ref{fig:More-qualitative}) and out-of-distribution (see the last two rows in Fig.~\ref{fig:More-qualitative}) cases. Notably, our method based on DreamBooth preserves the high fidelity and finer detail reconstruction benefits of vanilla DreamBooth, even in the context of blemished subject-driven generation.

\begin{figure*}[t]
  \centering    \includegraphics[width=0.9\textwidth, keepaspectratio]{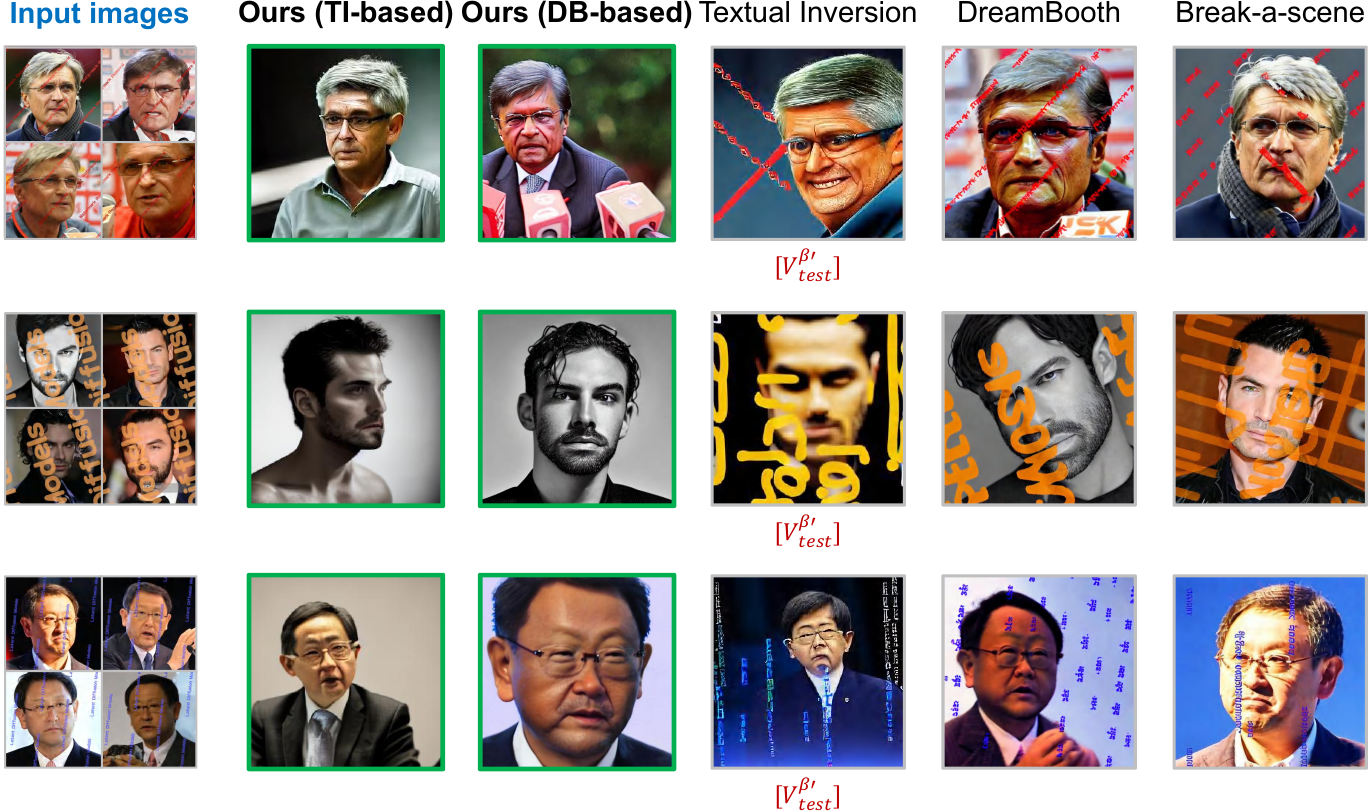}
      \caption{Additional qualitative comparisons - Human Faces.}
    \label{fig:More-qualitative_human}
\end{figure*}

In Fig.~\ref{fig:More-qualitative_human}, we show qualitative results for subjects with complex features (e.g., human faces) using our models, Textual Inversion, DreamBooth and Break-a-Scene~\cite{avrahami2023bas}. Break-a-Scene can separate multiple subjects inside one image. We use Break-a-scene to generate human-only images. However, we find that Break-a-scene fails to separate humans from artifacts, resulting in polluted images. 
As a result, our methods (\ie, TI-based and DB-based ArtiFade) consistently surpass Textual Inversion, DreamBooth, and Break-a-Scene, achieving high-quality image generation of complex data in in-distribution cases, as shown in the first two rows of Fig.~\ref{fig:More-qualitative_human}, and out-of-distribution cases, as illustrated in the last row of Fig.~\ref{fig:More-qualitative_human}.

\section{More Applications}
\begin{figure*}[t]
  \centering
    \includegraphics[width=0.85\textwidth, keepaspectratio]{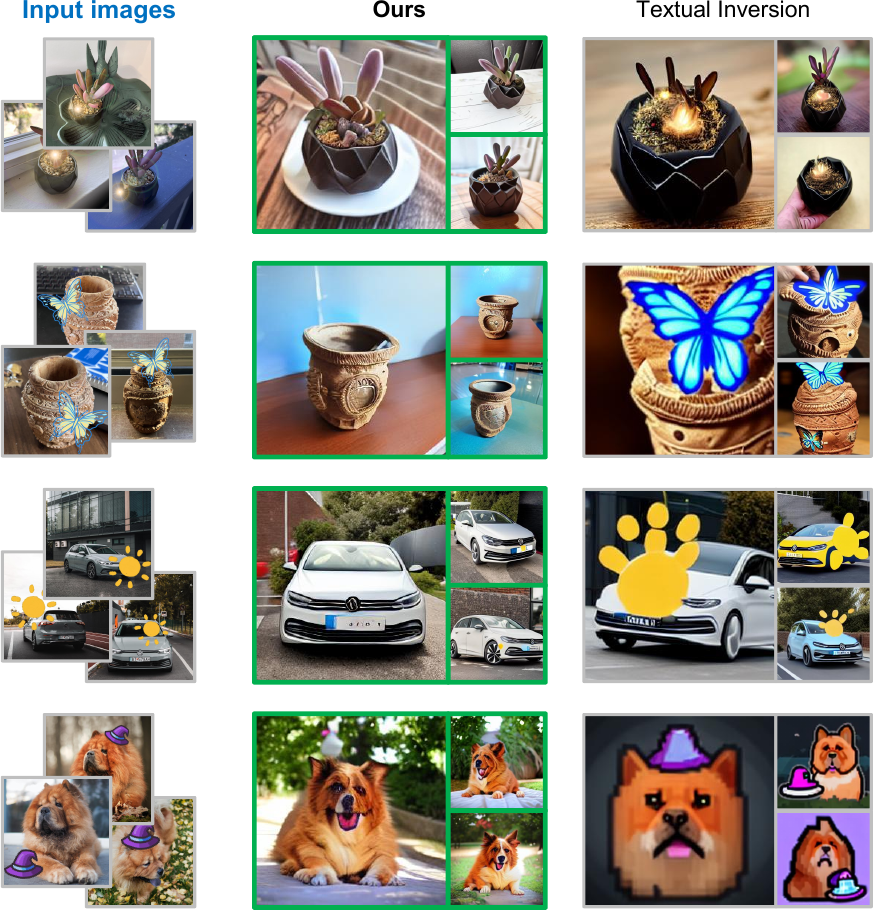}
      \caption{More applications. Our \WMModel can be used to eliminate various stickers and fix the incorrect image style.}
    \label{fig:More-applications}
\end{figure*}

We explore more applications of our \WMModel, demonstrating its versatility beyond watermark removal. As shown in Fig.~\ref{fig:More-applications}, our model exhibits the capability to effectively eliminate unwanted artifacts from images, enhancing their visual quality. Furthermore, our model showcases the ability to recover incorrect image styles induced by artifacts, thereby restoring the intended style of the images.

\section{Social Impact}
Our research addresses the emerging challenge of generating content from images with embedded watermarks, a scenario we term blemished subject-driven generation. Users often source images from the internet, some of which may contain watermarks intended to protect the original author's copyright and identity. However, our method is capable of removing various types of watermarks, potentially compromising the authorship and copyright protection. This could lead to increased instances of image piracy and the generation of illicit content. Hence, we advocate for legal compliance and the implementation of usage restrictions to govern the deployment of our technique and subsequent models in the future. \clearpage

\clearpage
\clearpage
\clearpage
\end{document}